\documentclass[final,5p,times,twocolumn]{elsarticle}
\usepackage[T1]{fontenc}
\usepackage[utf8]{inputenc}
\usepackage{amsmath,amssymb}
\usepackage{graphicx,booktabs,tabularx,array}
\usepackage{algorithm,algpseudocode}
\usepackage{microtype,needspace}
\usepackage[hidelinks]{hyperref}
\hypersetup{pdfauthor={Pengfeng Song, Yunkai Zhuang, Liye Shi},
  pdftitle={LMMR: Learned Multigranular Memory Retrieval for Small Language Model Agents}}
\usepackage{placeins,multicol,float}
\usepackage{adjustbox,multirow,dblfloatfix,caption}
\biboptions{numbers,sort&compress}
\newcolumntype{L}[1]{>{\raggedright\arraybackslash}p{#1}}
\newcolumntype{Y}{>{\raggedright\arraybackslash}X}
\makeatletter
\def\ps@pprintTitle{%
  \let\@oddhead\@empty
  \let\@evenhead\@empty
  \def\@oddfoot{\hfil\thepage\hfil}%
  \let\@evenfoot\@oddfoot}
\makeatother
\begin{document}
\begin{frontmatter}
\title{LMMR: Learned Multigranular Memory Retrieval for Small Language Model Agents}
\author[wanli]{Pengfeng Song\corref{cor1}}
\ead{sungkites@163.com}
\author[wanli]{Yunkai Zhuang}
\author[wanli]{Liye Shi}
\address[wanli]{School of Information and Intelligent Engineering, Zhejiang Wanli University, Ningbo 315100, China}
\cortext[cor1]{Corresponding author.}

\begin{abstract}
Small language model agents have limited space to read the memory needed for multi-hop questions. Relevant facts are often scattered across records whose surrounding text consumes much of that space. We propose LMMR, a learned multigranular memory retrieval method that selects both the next source and how much of it to read. LMMR links paragraphs and sentence fragments to their original sources and pairs views that supply the same missing support during training. Relevant distractors and repeated information provide contrasting examples. At inference, the reranker scores what each candidate adds to an anchor passage, and a selector chooses the evidence within a token budget. On 22,398 official validation questions from HotpotQA, 2WikiMultihopQA, and MuSiQue, a frozen 2B answer model reaches 61.47, 43.26, and 30.58 F1 under a 400-token budget. The gains over query expansion are 3.06, 2.14, and 4.55 points, while memory input decreases by 23.1--26.9\%. LMMR also uses 23.5--27.7\% fewer input tokens than a coarse retriever trained on the same questions. Against that retriever followed by LLMLingua-2 at comparable input lengths, F1 gains reach 3.99 points and are significant on HotpotQA and 2WikiMultihopQA after Holm correction. All eight small answer models evaluated on HotpotQA improve over query expansion and MMR in both F1 and exact match.
\end{abstract}
\end{frontmatter}

\section{Introduction}\label{sec:introduction}
Language-model agents use reasoning and feedback to guide their actions \cite{ref2,ref3} and retrieve stored experiences for later decisions \cite{ref1}. MemoryBank and MemGPT maintain information across context windows \cite{ref4,ref5}; A-MEM and Mem0 organize accumulated records through associations and updates \cite{ref6,ref7}. Two highly ranked memories can repeat a fact and leave a required relation elsewhere in the store. For a small answer model with a fixed input budget, the useful text may occupy only a few sentences of each retrieved record.

\begin{figure*}[!t]
\centering
\includegraphics[width=\linewidth]{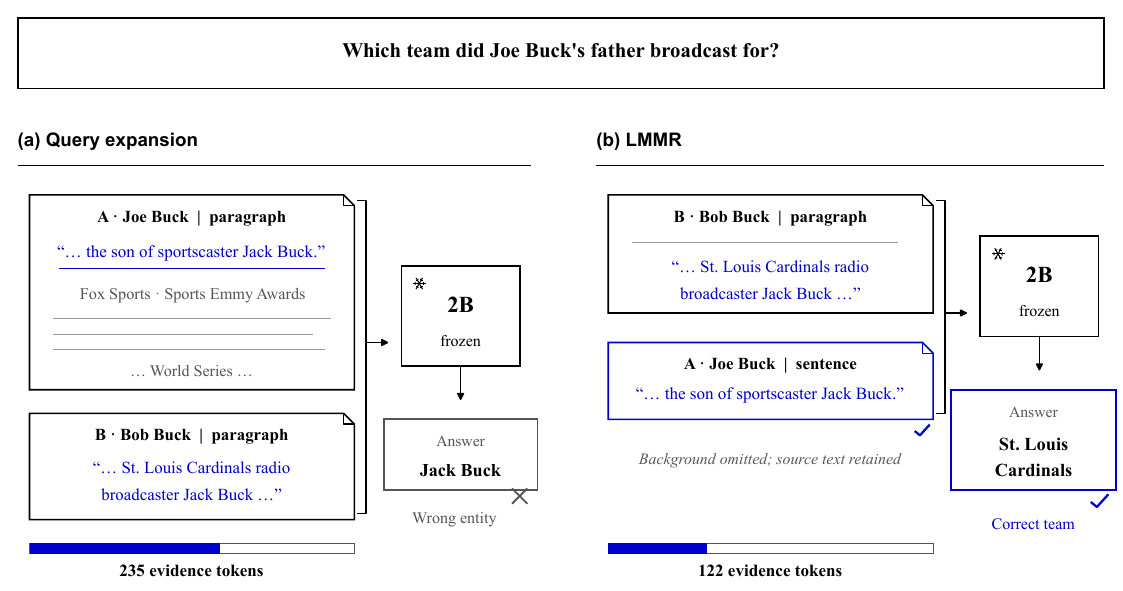}
\caption{The same question and source records yield different small-model answers. (a) Query expansion supplies two paragraphs (235 tokens); the 2B model returns the father's name. (b) LMMR uses the Bob Buck paragraph followed by a Joe Buck sentence (122 tokens), and the same model returns the correct team. Blue text marks supporting content; grey lines abbreviate background. Bars show evidence length on the same 400-token scale. Answers and lengths are from the recorded HotpotQA case.}\label{fig:concept}
\end{figure*}

Consider the question in Fig.~\ref{fig:concept}, drawn from HotpotQA \cite{ref48}: which team did Joe Buck's father broadcast for? Source A identifies Jack Buck as Joe's father. The second fact appears in a biography of Bob Buck, Jack's younger brother: Jack was a St.\ Louis Cardinals radio broadcaster. Both methods retrieve these two source records. Query expansion supplies 235 evidence tokens, but the 2B model answers with the father's name, ``Jack Buck''. LMMR changes the order and granularity, keeping the Bob Buck paragraph and one Joe Buck sentence. The resulting 122-token context gives the correct team. The shorter context retains the two relations needed to identify the team.

Evidence use depends on where the support appears in a long context \cite{ref8}; LongBench and RULER evaluate this capacity across long-input tasks \cite{ref9,ref11}. SimpleMem, LightMem, and small-model memory systems reduce storage or access costs \cite{ref14,ref15,ref16}. Even a relevant source can fill much of the input budget with background, leaving little room for a second relation.

Multi-hop retrieval already accounts for dependencies between sources. MDR conditions a later retrieval step on an earlier passage, and IRCoT uses intermediate reasoning to guide further search \cite{ref12,ref13}. IIER builds connections between chunks and retrieves evidence chains through a graph \cite{ref57}. MMR takes a different approach, penalizing overlap among relevant candidates \cite{ref40}. The amount of text retained from each source also affects how much evidence fits in the context. Some relations require paragraph-level context; others are stated in a single sentence.

TAG learns query-dependent weights over memory representations, MemGAS studies association across granularities, and GraP-Mem expands compact memories into their source contexts when more detail is needed \cite{ref25,ref24,ref26}. We use the correspondence between a paragraph and its fragments to train this choice. Views from the same source are paired when they supply the same missing support. The reranker learns to distinguish these views from relevant distractors and fragments that repeat known facts.

We introduce \emph{Learned Multigranular Memory Retrieval} (LMMR). Starting from an anchor paragraph, its reranker evaluates what another paragraph or fragment adds to the evidence, and a selector chooses a pair within the token budget. Training pairs coarse and fine views that keep the same missing support. The contributions are:
\begin{enumerate}
\item A memory representation that keeps paragraphs and sentence fragments linked to their original sources, allowing different amounts of context to be selected from different records.
\item A conditional reranker trained with equivalent coarse/fine support views, relevant non-supporting content, and repeated information. The learned score guides both source selection and granularity.
\item Full-validation evaluation on 22,398 multi-hop QA questions, with eight small answer models evaluated on HotpotQA. LMMR uses 23.5--27.7\% fewer tokens than training-matched coarse retrieval. At comparable input lengths, it gains 2.76 and 3.99 F1 over the matched retriever with LLMLingua-2 on HotpotQA and 2WikiMultihopQA, respectively; both differences remain significant after Holm correction.
\end{enumerate}

\section{Related work}\label{sec:related}
\subsection{Agent memory organization and learning}
MemoryBank and MemGPT address persistence and context management \cite{ref4,ref5}. A-MEM builds associations between stored records, and Mem0 consolidates and updates remembered facts \cite{ref6,ref7}. SAGE combines reflective feedback with a forgetting-curve-based memory mechanism \cite{ref56}. Learning has also been applied to memory operations: Memory-R1 and Mem-$\alpha$ use reinforcement learning \cite{ref18,ref19}, while Fine-Mem assigns finer-grained credit to those operations \cite{ref20}. MemAgent processes long contexts through segmented memory updates \cite{ref23}.

\subsection{Multigranular representations and selective access}
MemGAS links information across granularities \cite{ref24}. TAG learns a query-dependent mixture of chunks, atomic facts, triples, and summaries \cite{ref25}; GraP-Mem keeps compact records alongside source contexts and recovers detail when the evidence state is incomplete \cite{ref26}. In LMMR, paragraphs and fragments retain their original text and source mapping. A pair supplies a training example only when both views contain the same additional support under the current anchor, so the supervision specifies when the shorter view can replace its parent.

\subsection{Conditional retrieval and evidence completion}
RAG combines external retrieval with generation \cite{ref32}, using evidence that can be ranked by sparse scoring, dense retrieval, or late interaction \cite{ref36,ref33,ref34}. Multi-hop retrieval extends the search with information already found. MDR conditions a later retrieval step on an earlier passage, and IRCoT interleaves retrieval with intermediate reasoning \cite{ref12,ref13}. Chain-of-thought and least-to-most prompting also decompose reasoning into steps \cite{ref38,ref39}. IIER connects chunks through structural, keyword, and semantic relations to retrieve evidence chains \cite{ref57}.

MMR penalizes overlap with selected evidence \cite{ref40}. MemPursuit uses a query-reconstruction residual to guide the next selection \cite{ref28}, QEMem builds query-specific evidence sets \cite{ref29}, and ExecutionRAG tracks unresolved sub-questions \cite{ref30}. KITE executes retrieval plans over a structured memory index \cite{ref31}.

\subsection{Compact memory for small answer models}
SimpleMem compresses interaction history into indexed memory units and plans access according to the query \cite{ref14}. LightMem separates lightweight online processing from longer-term consolidation \cite{ref15}. Staged retrieval and reranking have also been used in small-model memory systems \cite{ref16}. LLMLingua-2 learns token retention from distilled compression data \cite{ref58}, shortening the text after retrieval. Qwen backbones and weight quantization make it practical to evaluate several small answer models under the same input constraints \cite{ref41,ref42}.

LoCoMo, LongMemEval, and MemoryAgentBench evaluate memory use over extended interactions \cite{ref45,ref46,ref47}. Multi-hop QA tests whether the retrieved text supplies the links needed to connect facts from different sources within a fixed token budget.

\section{Method}\label{sec:method}
\begin{figure*}[!t]
\centering\includegraphics[width=\linewidth]{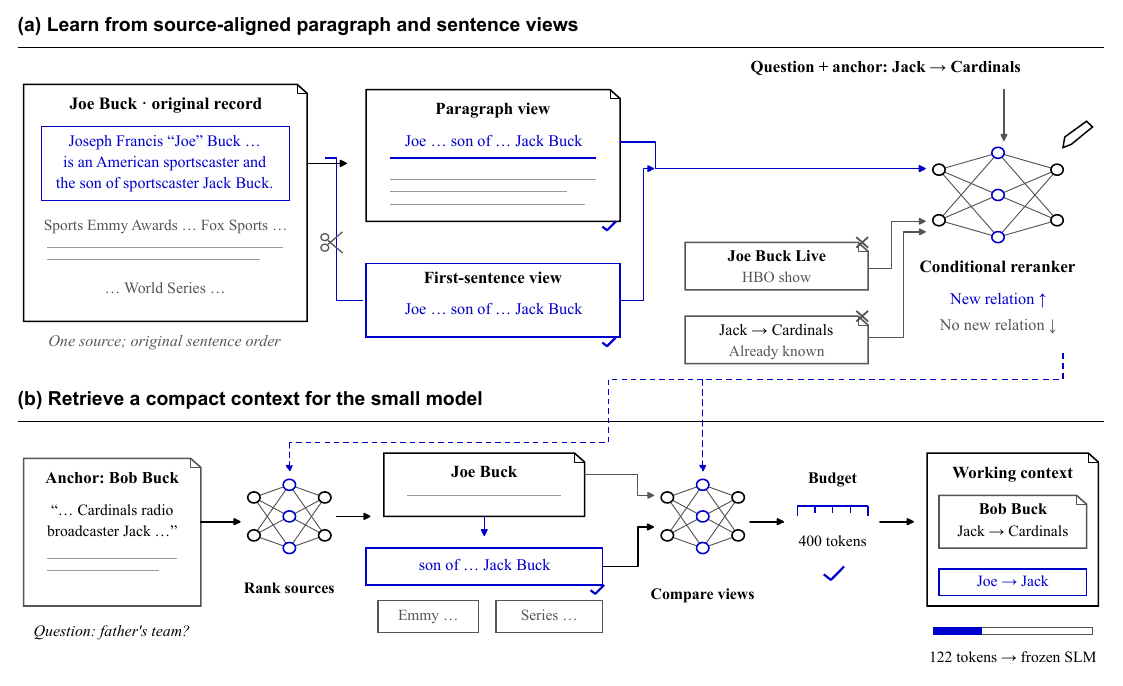}
\caption{LMMR illustrated through concrete memory views. (a) A source paragraph and its first-sentence view retain the same father relation. Under the same question and anchor, both supply new support; relevant non-supporting and redundant examples supervise the shared conditional reranker. (b) The reranker first ranks sources, then compares expanded sentence views with the paragraphs still available. Selection merges the anchor and one additional view within the token budget. Blue marks corresponding support, ellipses abbreviate displayed text, and dashed arrows show parameter sharing.}\label{fig:method}
\end{figure*}

LMMR starts with a paragraph that is relevant to the question, then selects another paragraph or fragment to complete the evidence. The scorer is trained to compare these candidates under the same anchor. Figure~\ref{fig:method} shows how the coarse/fine views enter training and retrieval.

\subsection{Problem definition and source-preserving representations}
For a question $q$ and memory collection $\mathcal{M}$, we select original-text evidence $E$ within a token budget $B$. Each paragraph node in $\mathcal{C}$ and fragment node in $\mathcal{F}$ is stored as $n=(p,g,I,x)$: source identifier $p$, granularity $g$, original sentence indices and character intervals $I$, and text $x$. The intervals link the two views of a source.
\begin{equation}
\begin{gathered}
E^*\subseteq\mathcal{C}\cup\mathcal{F},\qquad |E^*|\leq K_e,\\
T\!\left(\operatorname{Merge}(E^*)\right)\leq B.
\end{gathered}
\label{eq:budget}
\end{equation}
We set $K_e=2$, so each selection consists of an anchor and at most one additional node. The second node can be a paragraph or a fragment, with the budget shared freely between the two nodes. $\operatorname{Merge}$ combines overlapping intervals from the same source, restores sentence order, and adds source labels. $T$ counts the resulting tokens with the reference answer-model tokenizer.

Coarse nodes group adjacent sentences; fine nodes contain one sentence or two neighboring sentences. When a sentence exceeds the length limit, it is split into character intervals that keep the same source identifier. \ref{app:settings} gives the limits $\ell_C$ and $\ell_F$.

\subsection{Conditional scoring of additional information}
A frozen relevance score $s_0(q,c)$ ranks the initial coarse nodes. The next score depends on the question $q$, anchor $a$, and candidate $d$. We keep the base reranker's yes/no output form and compute the score from the final-token logits:
\begin{equation}
s_\theta(q,a,d)=z_{\mathrm{yes}}(q,a,d)-z_{\mathrm{no}}(q,a,d).
\label{eq:conditional}
\end{equation}
The input includes the anchor text and asks whether the candidate adds information still needed to answer the question. One scorer evaluates both granularities.

We also penalize lexical overlap with the anchor. Let $W(x)$ contain lowercase English words, excluding common stopwords, and adjacent word bigrams. The directional overlap $R(a,d)=|W(a)\cap W(d)|/(|W(d)|+\epsilon)$ measures how much of the candidate is already present. This term and the learned score enter the final selection objective.

\subsection{Source-consistent coarse-to-fine learning}
For each training question and anchor, let $G$ be the annotated support-sentence set and $U(n)$ the subset fully retained by node $n$. The missing support is $G_a=G\setminus U(a)$. A fine view $f$ and its positive coarse parent $c$ are paired when
\begin{equation}
f\subseteq c,\qquad
U(f)\cap G_a=U(c)\cap G_a\ne\varnothing.
\label{eq:positives}
\end{equation}
For example, with source A as the anchor in Fig.~\ref{fig:concept}, a view from source B must keep the support linking Jack Buck to his broadcasting affiliation. A fragment that merely mentions his name cannot replace the coarse positive. We choose the shortest fine view satisfying Eq.~\eqref{eq:positives}. A group without a qualifying coarse--fine pair uses an available supported view of the source.

Hard negatives are high-scoring non-supporting candidates mined with frozen query expansion at each granularity. The anchor and its fine views serve as redundant negatives. Write their scores as $s_v^-$ and $s_v^{\mathrm{red}}$, with $s_v^+$ for a positive view $v$. For the positive-view group $V^+$, the ranking loss is
\begin{equation}
\begin{split}
\mathcal{L}_{\mathrm{rank}}=\frac{1}{|V^+|}\sum_{v\in V^+}
\big[&\operatorname{sp}(s_v^- -s_v^+ +m)\\
&+\lambda_r\operatorname{sp}(s_v^{\mathrm{red}}-s_v^+ +m)\big],
\end{split}
\label{eq:rankloss}
\end{equation}
where $\operatorname{sp}(t)=\log(1+\exp(t))$, $m$ is the ranking margin, and $\lambda_r$ weights redundant negatives. Binary supervision is applied to positive and hard-negative scores:
\begin{equation}
\begin{aligned}
\mathcal{L}_B=-\frac{1}{2|V^+|}\sum_{v\in V^+}
\big[&\log\sigma(s_v^+)\\
&+\log\left(1-\sigma(s_v^-)\right)\big].
\end{aligned}
\label{eq:bceloss}
\end{equation}
When both coarse and fine positives exist, an auxiliary term brings their sigmoid scores closer:
\begin{equation}
\mathcal{L}_c=\operatorname{SmoothL1}\left(\sigma(s_c^+),\sigma(s_f^+)\right).
\label{eq:consistency}
\end{equation}
It is zero for a single-granularity positive group. The overall objective is
\begin{equation}
\mathcal{L}=\mathcal{L}_{\mathrm{rank}}+\lambda_B\mathcal{L}_B+\lambda_c\mathcal{L}_c.
\label{eq:total}
\end{equation}
The same-source support pairing determines the positive groups used by the ranking and binary losses; $\mathcal{L}_c$ adds score smoothing. We average losses by question and adapt Qwen3-Reranker 0.6B \cite{ref51} using LoRA \cite{ref52} on the attention query and value projections. All other base parameters stay frozen. Training settings are listed in \ref{app:settings}.

\subsection{Budgeted conditional hierarchical selection}
We keep at most $K_a$ coarse anchors from $s_0$. If the leading score exceeds the runner-up by more than $\tau$, only the first anchor is used. For each anchor $a$, $s_\theta$ scores the coarse candidates not fully contained in $a$. Fine views are then expanded from the top $K_s$ distinct parent sources $P_a$:
\begin{equation}
\begin{split}
\mathcal{H}(a)={}&\{c\in\mathcal{C}:c\nsubseteq a\}\\
&\cup\{f\in\mathcal{F}:\operatorname{src}(f)\in P_a,\ f\nsubseteq a\}.
\end{split}
\label{eq:hierarchy}
\end{equation}
Containment is checked against the original intervals. A fragment from the anchor's source can still be selected if it covers new text. We keep the coarse candidates as well and choose
\begin{equation}
\begin{aligned}
(a^*,d^*)&=\underset{a\in\mathcal{A},\ d\in\mathcal{H}(a)}{\arg\max}\\
&\quad\left[s_0(q,a)+\alpha s_\theta(q,a,d)-\beta R(a,d)\right],
\end{aligned}
\label{eq:selection}
\end{equation}
subject to $T(\operatorname{Merge}(\{a,d\}))\leq B$ and the reference-prompt limit. The weights $\alpha$ and $\beta$ balance the learned score and overlap penalty. If no pair fits, the leading anchor is returned. Otherwise, the merged intervals form the answer model's input. Algorithm~\ref{alg:lmmr} summarizes the procedure.

\begin{algorithm}[!tbp]
\caption{Budgeted multigranular memory retrieval}\label{alg:lmmr}
\footnotesize
\begin{algorithmic}[1]
\Require Question $q$, memories $\mathcal M$, budget $B$
\Require Scorers $s_0,s_\theta$; $K_a,K_s,\tau,\alpha,\beta,L_{\max}$
\Ensure Merged evidence $e^*$
\State $\mathcal C,\mathcal F\gets\operatorname{BuildNodes}(\mathcal M)$
\State $\mathcal A\gets\operatorname{TopK}_{K_a}(\mathcal C;s_0)$
\State $a_1,a_2\gets$ leading anchors, if present
\If{$|\mathcal A|>1$ and $s_0(q,a_1)-s_0(q,a_2)>\tau$}
\State $\mathcal A\gets\{a_1\}$
\EndIf
\State $\mathrm{best}\gets-\infty$; $E^*\gets\{a_1\}$
\For{each $a\in\mathcal A$}
\State $\mathcal C_a\gets\{c\in\mathcal C:c\nsubseteq a\}$
\State Score $\mathcal C_a$ with $s_\theta(q,a,\cdot)$
\State $P_a\gets$ top $K_s$ distinct sources
\State Build $\mathcal H(a)$ by Eq.~\eqref{eq:hierarchy}
\State Score the expanded fine candidates
\For{each $d\in\mathcal H(a)$}
\State $e\gets\operatorname{Merge}(\{a,d\})$
\If{$T(e)\leq B$ and prompt length $\leq L_{\max}$}
\State $J\gets s_0(q,a)+\alpha s_\theta(q,a,d)-\beta R(a,d)$
\If{$J>\mathrm{best}$}
\State $(\mathrm{best},E^*)\gets(J,\{a,d\})$
\EndIf
\EndIf
\EndFor
\EndFor
\State \Return $\operatorname{Merge}(E^*)$
\end{algorithmic}
\end{algorithm}

With $F_s$ expanded fine nodes per anchor, scoring costs $O(|\mathcal{C}|+K_a(|\mathcal{C}|+F_s))$ operations. Coarse scores are reused; the answer model runs once after selection.

\onecolumn
\begin{multicols}{2}
\raggedcolumns
\section{Experiments}\label{sec:experiments}
\subsection{Experimental setting}
We evaluate 7,405, 12,576, and 2,417 official validation questions in HotpotQA, 2WikiMultihopQA, and MuSiQue-Ans, respectively \cite{ref48,ref49,ref50}. The main setting uses Qwen3.5-2B and a 400-token evidence budget. Answer-model weights remain unchanged in all comparisons, including the eight-model evaluation. HotpotQA uses a 30-paragraph candidate pool; the external datasets retain their original 10 and 20 paragraphs. Additional HotpotQA distractors are retrieved by BM25 using the question and initial anchor. LMMR is trained on 1,600 HotpotQA training questions.

We compare query expansion (QE), normalized MMR, coarse-only weak-negative adaptation, frozen hierarchical selection (FH), and frozen conditional hierarchical selection (FCH). The additional controls are a training-matched coarse-only retriever, QE followed by LLMLingua-2, and the matched retriever followed by LLMLingua-2 \cite{ref58}. The matched retriever shares the training questions and optimization settings of LMMR; its selector considers coarse candidates under the same anchor and budget rules. The hierarchical controls share LMMR's candidate construction and budget constraints. We report answer F1 and exact match (EM), together with evidence length and support coverage. Paired bootstrap intervals \cite{ref53}, model configurations, and metric definitions are summarized in \ref{app:settings}.

\subsection{Answer quality and input compression}
\begin{figure}[H]
\centering\includegraphics[width=.90\linewidth]{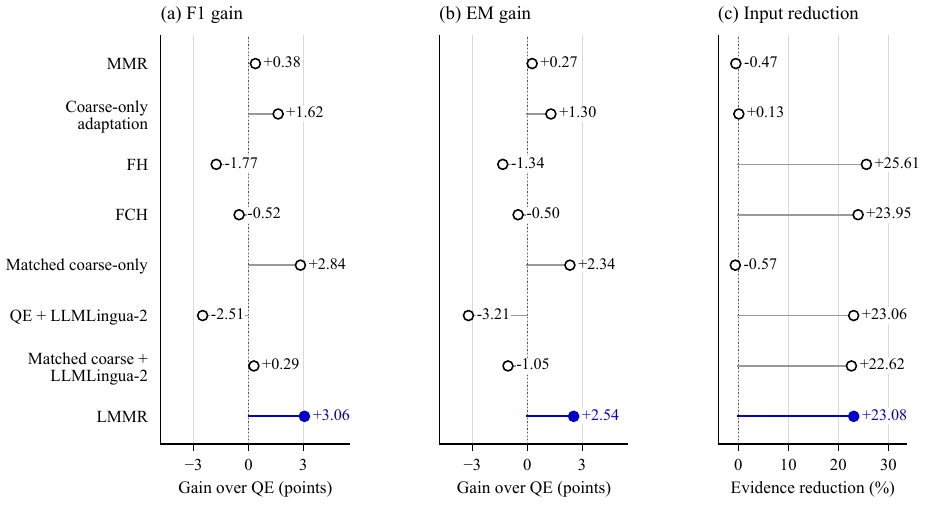}
\caption{Answer-quality differences and evidence-input reduction relative to QE on HotpotQA, including the training-matched and learned-compression controls. F1 (a) and EM (b) use identical axis ranges. Panel (c) shows the reduction in mean evidence tokens. All settings use 7,405 questions; differences are calculated before rounding.}\label{fig:main}
\end{figure}

On HotpotQA, LMMR reaches 61.47 F1 and 45.96 EM. The gains over QE are 3.06 F1 and 2.54 EM; the F1 gain has a 95\% interval of $[2.43,3.70]$. LMMR also leads MMR by 2.68 F1. It uses 185.32 evidence tokens on average, down from 240.92 for QE (23.1\%; Table~\ref{tab:strongbaselines}; Fig.~\ref{fig:main}).
\end{multicols}

\begin{table}[H]
\centering
\caption{Training-matched retrieval and learned compression on the full validation sets. All methods use the 2B answer model and a 400-token evidence budget.}\label{tab:strongbaselines}
\small
\setlength{\tabcolsep}{3pt}
\renewcommand{\arraystretch}{1.13}
\begin{tabularx}{\linewidth}{@{}Yrrrrrrrrr@{}}
\toprule
\multirow{2}{*}{Method} & \multicolumn{3}{c}{HotpotQA (7,405)} & \multicolumn{3}{c}{2Wiki (12,576)} & \multicolumn{3}{c}{MuSiQue (2,417)} \\
\cmidrule(lr){2-4}\cmidrule(lr){5-7}\cmidrule(lr){8-10}
& F1 $\uparrow$ & EM $\uparrow$ & Tokens $\downarrow$ & F1 $\uparrow$ & EM $\uparrow$ & Tokens $\downarrow$ & F1 $\uparrow$ & EM $\uparrow$ & Tokens $\downarrow$ \\
\midrule
Query expansion & 58.42 & 43.42 & 240.92 & 41.12 & 31.67 & 242.47 & 26.02 & 17.17 & 259.01 \\
Normalized MMR & 58.79 & 43.69 & 242.05 & 41.49 & 31.99 & 243.72 & 25.97 & 17.13 & 259.96 \\
Matched coarse-only & 61.25 & 45.75 & 242.30 & 42.71 & \textbf{32.82} & 245.97 & 28.50 & 19.07 & 260.77 \\
QE + LLMLingua-2 & 55.90 & 40.20 & 185.35 & 37.54 & 26.06 & 186.38 & 26.39 & 16.80 & 200.18 \\
Matched coarse + LLMLingua-2 & 58.71 & 42.36 & 186.42 & 39.27 & 27.59 & 189.05 & 29.22 & 18.87 & 201.32 \\
\textbf{LMMR} & \textbf{61.47} & \textbf{45.96} & \textbf{185.32} & \textbf{43.26} & 32.63 & \textbf{177.94} & \textbf{30.58} & \textbf{20.27} & \textbf{189.21} \\
\bottomrule
\end{tabularx}
\end{table}

\begin{multicols}{2}
\raggedcolumns
\subsection{Training-matched comparisons and loss ablations}
On HotpotQA, the matched coarse-only retriever reaches 61.25 F1 with 242.30 mean evidence tokens. LMMR scores 61.47 F1 using 185.32 tokens, a 23.5\% reduction. Adding LLMLingua-2 to the matched retriever gives 58.71 F1 and 42.36 EM with 186.42 tokens (Table~\ref{tab:strongbaselines}). Relative to the matched retriever followed by LLMLingua-2, LMMR gains 2.76, 3.99, and 1.35 F1 points on HotpotQA, 2WikiMultihopQA, and MuSiQue, respectively. The HotpotQA and 2Wiki differences remain significant after Holm correction (Table~\ref{tab:statistics}).
\end{multicols}

\begin{table}[H]
\centering
\caption{Full-validation ablation of training and loss components on HotpotQA.}\label{tab:ablation}
\small
\setlength{\tabcolsep}{3.5pt}
\renewcommand{\arraystretch}{1.13}
\begin{tabularx}{\linewidth}{@{}Yccrrrrr@{}}
\toprule
\multirow{2}{*}{Configuration} & \multicolumn{2}{c}{Learning} & \multicolumn{2}{c}{Answer quality} & \multicolumn{3}{c}{Evidence} \\
\cmidrule(lr){2-3}\cmidrule(lr){4-5}\cmidrule(lr){6-8}
& Adapted & Paired views & F1 $\uparrow$ & EM $\uparrow$ & Complete $\uparrow$ & Density $\uparrow$ & Tokens $\downarrow$ \\
\midrule
FH & -- & -- & 56.65 & 42.08 & 63.46 & 42.25 & \textbf{179.22} \\
FCH & -- & -- & 57.90 & 42.92 & 65.91 & 42.13 & 183.22 \\
\addlinespace[3pt]
Coarse-only adaptation & $\checkmark$ & -- & 60.04 & 44.71 & 80.85 & 33.90 & 240.61 \\
Without score smoothing & $\checkmark$ & $\checkmark$ & \textbf{61.49} & 45.86 & 73.09 & 43.76 & 185.16 \\
Without redundant negatives & $\checkmark$ & $\checkmark$ & 61.23 & 45.66 & 72.90 & \textbf{43.78} & 185.54 \\
\addlinespace[3pt]
Matched coarse-only & $\checkmark$ & -- & 61.25 & 45.75 & \textbf{83.08} & 34.27 & 242.30 \\
\textbf{LMMR} & $\checkmark$ & $\checkmark$ & 61.47 & \textbf{45.96} & 72.99 & 43.75 & 185.32 \\
\bottomrule
\end{tabularx}
\end{table}

\begin{multicols}{2}
\raggedcolumns
With the hierarchical candidate structure held fixed, source-consistent adaptation raises F1 from 57.90 to 61.47 and EM from 42.92 to 45.96 relative to FCH. Removing redundant-negative supervision reduces F1 and EM by 0.24 and 0.30 points; removing the auxiliary score-smoothing term leaves F1 essentially unchanged (Table~\ref{tab:ablation}).

\subsection{Candidate-pool interference}
\begin{figure}[H]
\centering\includegraphics[width=0.97415\linewidth]{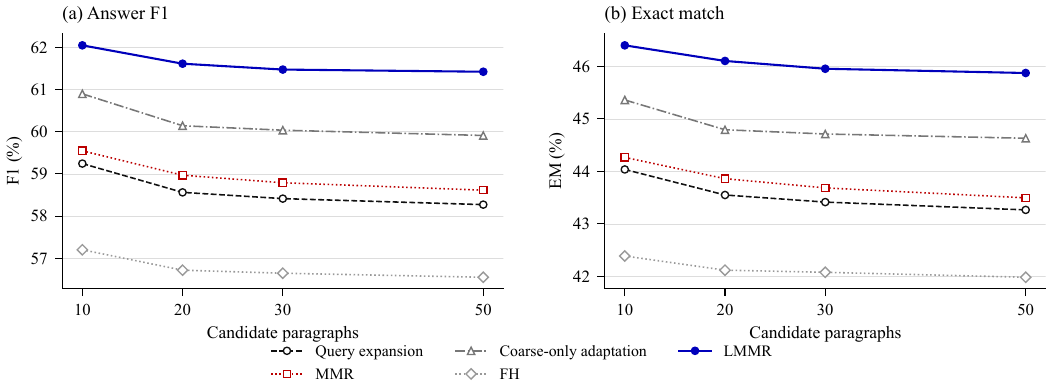}
\caption{Answer F1 and EM as the HotpotQA candidate pool increases from 10 to 50 paragraphs. Each setting uses 7,405 validation questions.}\label{fig:interference}
\end{figure}

As the pool grows from 10 to 50 paragraphs, QE decreases from 59.24 to 58.27 F1, whereas LMMR decreases from 62.05 to 61.42. LMMR loses 0.63 points to QE's 0.97, increasing its lead from 2.80 to 3.15 F1 (Fig.~\ref{fig:interference}; Table~\ref{tab:pool}).
\end{multicols}

\begin{multicols}{2}
\raggedcolumns
\subsection{Score weight and evidence budget}
\begin{figure}[H]
\centering\includegraphics[width=0.97849\linewidth]{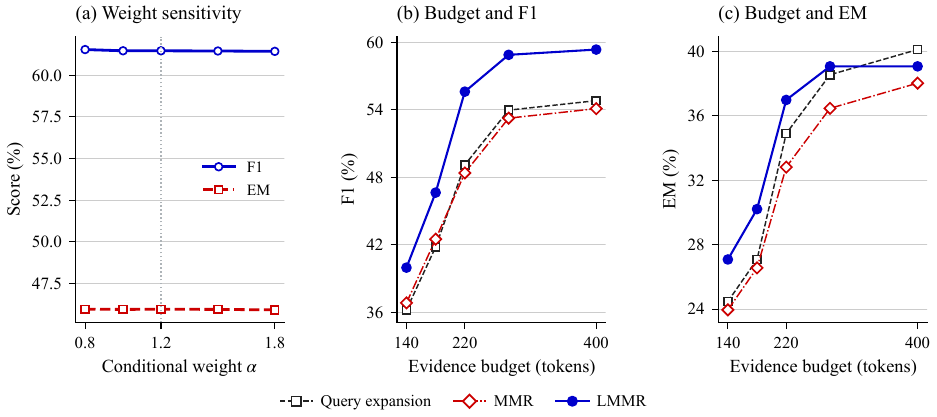}
\caption{Conditional-score weight and evidence-budget analyses. (a) Weight sensitivity on 7,405 HotpotQA validation questions. (b,c) Fixed-evidence clipping on 192 questions.}\label{fig:parameters}
\end{figure}

Changing $\alpha$ from 0.8 to 1.8 yields 61.44--61.55 F1, 45.91--45.96 EM, and 185.00--185.95 mean evidence tokens. Performance changes little over this range (Fig.~\ref{fig:parameters}a; Table~\ref{tab:alpha}).

On the 192-question fixed-evidence clipping subset, reducing the budget from 400 to 220 tokens lowers QE from 54.83 to 49.12 F1 and LMMR from 59.37 to 55.63. The gain increases from 4.55 to 6.51 points. LMMR remains ahead by 4.84 and 3.76 F1 at budgets of 180 and 140 tokens, respectively (Fig.~\ref{fig:parameters}b,c; Table~\ref{tab:budget}). The largest gain in this sweep occurs at 220 tokens, before the most severe clipping.
\end{multicols}

\begin{multicols}{2}
\raggedcolumns
\subsection{Transfer across small answer models}
\begin{figure}[H]
\centering\includegraphics[width=.90\linewidth]{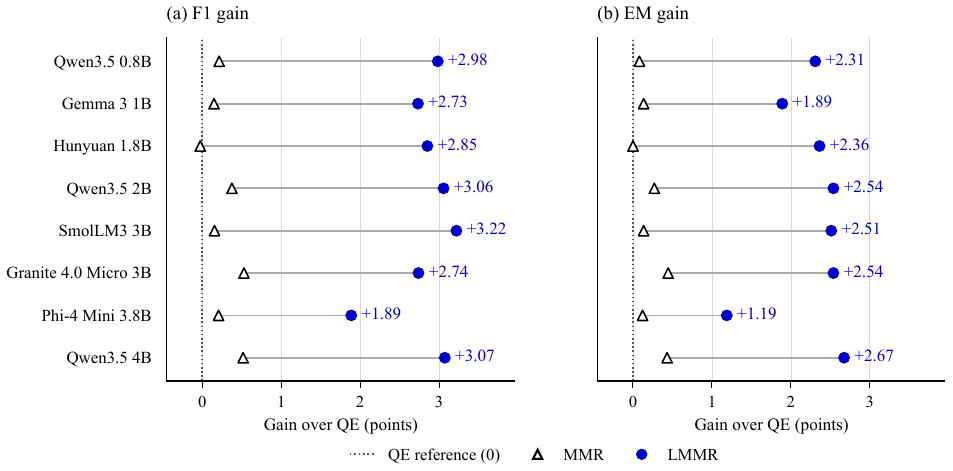}
\caption{F1 and EM gains over QE for eight small answer models on 7,405 HotpotQA validation questions. Both panels use the same gain scale; the dotted line marks the QE reference at zero. Triangles denote MMR and blue circles denote LMMR. Labels give the LMMR gains, calculated before rounding. Absolute scores are reported in Table~\ref{tab:models}.}\label{fig:models}
\end{figure}

LMMR exceeds QE and MMR in both F1 and EM for each of the eight answer models. Relative to QE, F1 improves by 1.89--3.22 points and EM by 1.19--2.67 points. Qwen3.5-4B reaches 71.76 F1 and 57.18 EM (Table~\ref{tab:models}; Fig.~\ref{fig:models}). These models range from 0.8B to 4B; all use the same trained retrieval model. Additional answer-model scores are reported in Table~\ref{tab:nemotron}.
\end{multicols}

\begin{table}[H]
\centering
\caption{Eight small answer models on 7,405 HotpotQA validation questions.}\label{tab:models}
\small
\setlength{\tabcolsep}{3.5pt}
\renewcommand{\arraystretch}{1.13}
\begin{tabularx}{\linewidth}{@{}Yrrrrrr@{}}
\toprule
\multirow{2}{*}{Answer model} & \multicolumn{2}{c}{Query expansion} & \multicolumn{2}{c}{MMR} & \multicolumn{2}{c}{LMMR} \\
\cmidrule(lr){2-3}\cmidrule(lr){4-5}\cmidrule(lr){6-7}
& F1 $\uparrow$ & EM $\uparrow$ & F1 $\uparrow$ & EM $\uparrow$ & F1 $\uparrow$ & EM $\uparrow$ \\
\midrule
Qwen3.5 0.8B & 46.19 & 33.48 & 46.41 & 33.56 & \textbf{49.18} & \textbf{35.79} \\
Gemma 3 1B & 42.70 & 29.76 & 42.85 & 29.90 & \textbf{45.43} & \textbf{31.65} \\
Hunyuan 1.8B & 39.36 & 27.75 & 39.33 & 27.75 & \textbf{42.21} & \textbf{30.11} \\
Qwen3.5 2B & 58.42 & 43.42 & 58.79 & 43.69 & \textbf{61.47} & \textbf{45.96} \\
SmolLM3 3B & 37.27 & 18.53 & 37.43 & 18.66 & \textbf{40.49} & \textbf{21.04} \\
Granite 4.0 Micro 3B & 65.64 & 51.24 & 66.17 & 51.68 & \textbf{68.38} & \textbf{53.77} \\
Phi-4 Mini 3.8B & 61.24 & 45.40 & 61.45 & 45.52 & \textbf{63.13} & \textbf{46.59} \\
Qwen3.5 4B & 68.69 & 54.50 & 69.21 & 54.94 & \textbf{71.76} & \textbf{57.18} \\
\bottomrule
\end{tabularx}
\end{table}

\begin{multicols}{2}
\raggedcolumns
\subsection{Cross-dataset and question-type results}

The HotpotQA-trained scorer reaches 43.26 F1 and 32.63 EM on 2WikiMultihopQA, ahead of QE and MMR by 2.14 and 1.77 F1, with 26.6\% less evidence input. On MuSiQue, the scores are 30.58 F1 and 20.27 EM; the F1 gains are 4.55 over QE and 4.61 over MMR, and input decreases by 26.9\% (Tables~\ref{tab:strongbaselines} and~\ref{tab:external}). Table~\ref{tab:aliases} gives alias-aware MuSiQue scores in addition to the single-reference scores used here.

\begin{figure}[H]
\centering\includegraphics[width=0.96832\linewidth]{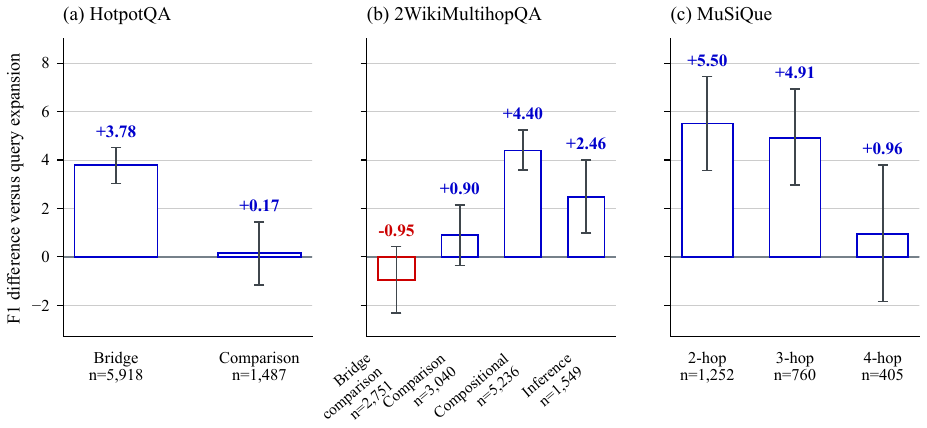}
\caption{Question-type F1 gains over QE, with pointwise 95\% paired-bootstrap intervals. The number of questions in each group is denoted by $n$.}\label{fig:types}
\end{figure}

The larger question-type gains concern cross-source composition: 3.78 F1 on HotpotQA bridge questions, 4.40 on 2Wiki compositional questions, and 5.50 and 4.91 on MuSiQue two-hop and three-hop questions (Fig.~\ref{fig:types}; Table~\ref{tab:types}). \ref{app:published} lists published results with their retrieval settings.
\end{multicols}

\begin{multicols}{2}
\raggedcolumns
\subsection{Support density and error distribution}\label{sec:mechanism}
Most of the removed tokens lie outside the annotated support (Table~\ref{tab:main}). Mean annotated support text changes from 75.35 to 74.54 tokens, while other input falls from 165.57 to 110.78 tokens, a 33.1\% reduction. Mean per-question support density increases from 33.47\% to 43.75\%, support-source recall from 90.07\% to 93.46\%, and non-boolean answer-in-context rate (AIC) from 89.66\% to 91.48\%.
\end{multicols}

\begin{table}[H]
\centering
\caption{Answer and evidence quality on 7,405 HotpotQA validation questions. Support-source recall, support completeness, support density, and AIC are percentages.}\label{tab:main}
\small
\setlength{\tabcolsep}{3.5pt}
\renewcommand{\arraystretch}{1.13}
\begin{tabularx}{\linewidth}{@{}Yrrrrrrr@{}}
\toprule
\multirow{2}{*}{Method} & \multicolumn{2}{c}{Answer quality} & \multicolumn{2}{c}{Support coverage (\%)} & \multicolumn{2}{c}{Evidence quality (\%)} & \multirow{2}{*}{Tokens $\downarrow$} \\
\cmidrule(lr){2-3}\cmidrule(lr){4-5}\cmidrule(lr){6-7}
& F1 $\uparrow$ & EM $\uparrow$ & Source $\uparrow$ & Complete $\uparrow$ & Density $\uparrow$ & AIC $\uparrow$ & \\
\midrule
Query expansion & 58.42 & 43.42 & 90.07 & 77.87 & 33.47 & 89.66 & 240.92 \\
Normalized MMR & 58.79 & 43.69 & 90.43 & 78.18 & 33.46 & 90.33 & 242.05 \\
Coarse-only adaptation & 60.04 & 44.71 & 91.53 & 80.85 & 33.90 & 92.23 & 240.61 \\
FH & 56.65 & 42.08 & 89.50 & 63.46 & 42.25 & 84.87 & \textbf{179.22} \\
FCH & 57.90 & 42.92 & 89.57 & 65.91 & 42.13 & 86.09 & 183.22 \\
\addlinespace[3pt]
Matched coarse-only & 61.25 & 45.75 & 92.65 & \textbf{83.08} & 34.27 & \textbf{93.44} & 242.30 \\
\textbf{LMMR} & \textbf{61.47} & \textbf{45.96} & \textbf{93.46} & 72.99 & \textbf{43.75} & 91.48 & 185.32 \\
\bottomrule
\end{tabularx}
\end{table}

\begin{multicols}{2}
\raggedcolumns
Strict support completeness requires every annotated support sentence for a question to be fully retained. \emph{Span miss} identifies questions whose answers fail EM despite a supporting anchor and coverage of all supporting sources, with incomplete support-sentence retention. A 96-token fine node may keep a local fact without keeping every annotated sentence in its paragraph. LMMR has lower support completeness (77.87\% to 72.99\%) and a larger span-miss category (2.57\% to 9.59\%).

\begin{figure}[H]
\centering\includegraphics[width=0.95359\linewidth]{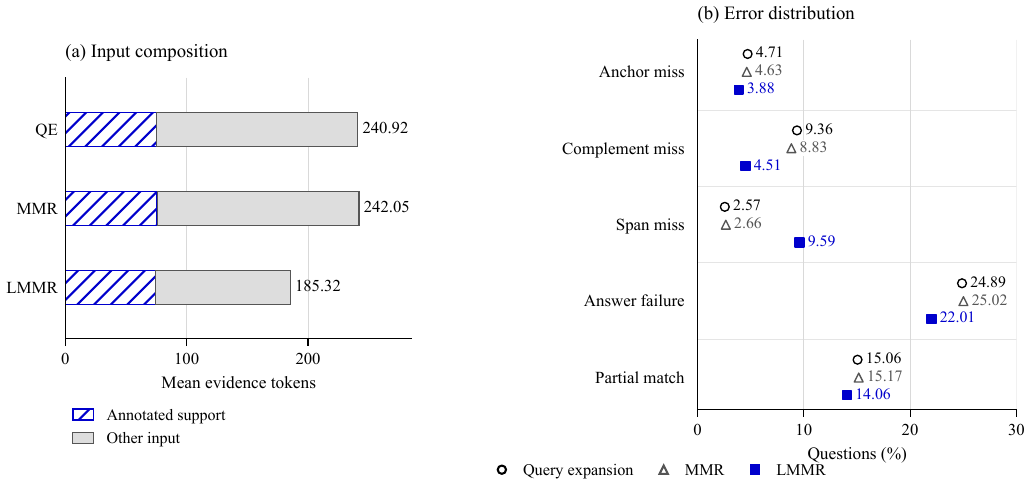}
\caption{Mean evidence composition (a) and five mutually exclusive error categories (b) on HotpotQA. The three methods occupy separate tracks within each error category. Point labels are percentages of 7,405 validation questions. QE denotes query expansion.}\label{fig:diagnostics}
\end{figure}

Missing-complement-source errors fall from 9.36\% to 4.51\% and anchor misses from 4.71\% to 3.88\%. The answer-failure category decreases from 24.89\% to 22.01\% (Fig.~\ref{fig:diagnostics}).
\end{multicols}

\begin{multicols}{2}
\raggedcolumns
\subsection{Question answering on Jetson Orin}
On 192 questions with retrieval on an A100 and answer generation on Jetson Orin, evidence input decreases from 232.34 to 182.18 tokens, or 21.6\%. Orin prompt encoding falls from 528.51 to 464.83\,ms, a 12.0\% reduction. Generation throughput is 19.78 tokens/s, and peak process resident memory is 9.32\,GiB. Total request latency is 2.27\,s for LMMR and 1.93\,s for QE (Table~\ref{tab:edge}).
\end{multicols}

\begin{table}[H]
\centering
\caption{Measured execution on 192 questions: evidence selection on A100 and answering on Jetson Orin.}\label{tab:edge}
\small
\setlength{\tabcolsep}{3.5pt}
\renewcommand{\arraystretch}{1.13}
\begin{tabularx}{\linewidth}{@{}Yrrr@{}}
\toprule
Metric & QE & FCH & LMMR \\
\midrule
Mean evidence tokens & 232.34 & \textbf{178.01} & 182.18 \\
Evidence selection, A100 (ms) & \textbf{690.81} & 1116.36 & 1081.23 \\
Prompt encoding, Orin (ms) & 528.51 & \textbf{456.03} & 464.83 \\
Answer generation, Orin (ms) & 360.96 & \textbf{360.17} & 367.21 \\
Total request latency (ms) & \textbf{1927.87} & 2291.09 & 2271.93 \\
Generation throughput, Orin (tokens/s) & 19.68 & 19.73 & \textbf{19.78} \\
\bottomrule
\end{tabularx}
\end{table}

\begin{figure}[H]
\centering\includegraphics[width=\linewidth]{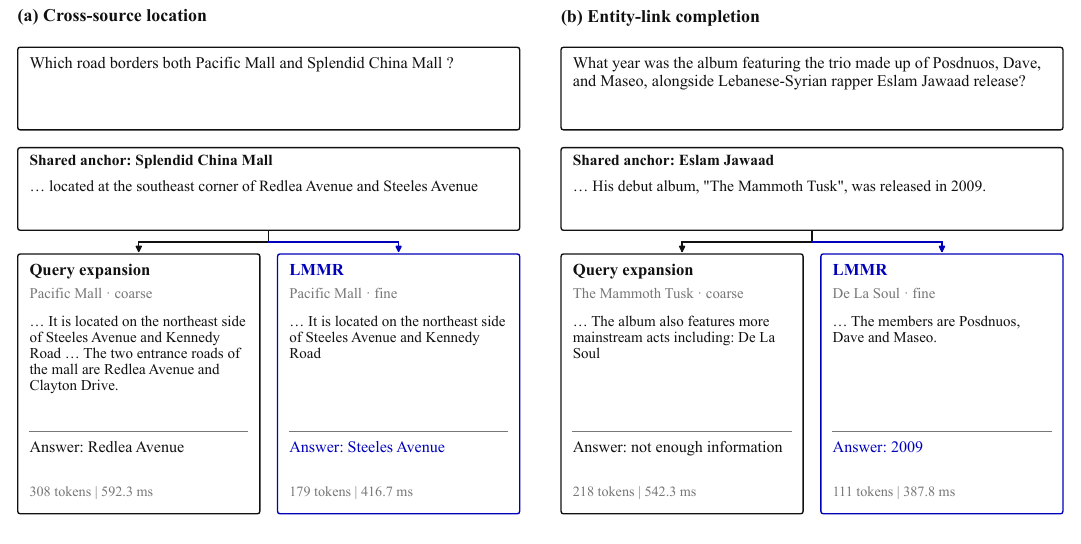}
\caption{Recorded question-answering examples on Jetson Orin. The methods share a coarse anchor but select a full paragraph or a fine fragment as subsequent memory. Parallel evidence cards show original excerpts and recorded model answers. The lower annotations report evidence length and Orin prompt-encoding time.}\label{fig:cases}
\end{figure}

\begin{multicols}{2}
\raggedcolumns
Figure~\ref{fig:cases} shows two answers generated on Jetson Orin. In the road-intersection example, evidence decreases from 308 to 179 tokens and the answer changes from ``Redlea Avenue'' to the annotated ``Steeles Avenue.'' In the album-year example, input decreases from 218 to 111 tokens and the answer changes from ``not enough information'' to ``2009.'' The input reductions are 41.9\% and 49.1\%, respectively.
\end{multicols}

\begin{multicols}{2}
\raggedcolumns
\section{Discussion}\label{sec:discussion}
LMMR uses 23.5\% less input than the matched coarse-only retriever at comparable HotpotQA answer quality. When that retriever's output is compressed with LLMLingua-2 to a similar length, LMMR leads by 2.76 F1 (Table~\ref{tab:strongbaselines}). The training pairs link a fine view to its parent when both supply the same missing relation. The selector can then keep that relation in its original wording and discard surrounding text.

The gains on bridge and compositional questions fit this use of granularity. An anchor establishes one relation; the next view supplies the link needed to reach the answer. A paragraph can retain context spanning several sentences, while a fine view makes room when the link is stated locally. The same learned retriever improves eight frozen answer models over QE and MMR, so the benefit carries across answer-model sizes and architectures.

\section{Conclusion}\label{sec:conclusion}
LMMR learns to select paragraphs and sentence fragments according to the support they add to previously retrieved memory. On 22,398 multi-hop QA questions, it improves F1 by 2.14--4.55 points over QE while reducing input by 23.1--26.9\%. It also uses 23.5--27.7\% fewer tokens than a training-matched coarse retriever. At comparable input lengths, LMMR gains 2.76 and 3.99 F1 over that retriever followed by LLMLingua-2 on HotpotQA and 2WikiMultihopQA; both differences remain significant after Holm correction. A single trained retriever benefits all eight small answer models evaluated.

\section*{CRediT authorship contribution statement}
\textbf{Pengfeng Song:} Conceptualization, Methodology, Software, Validation, Writing -- original draft, Writing -- review \& editing, Funding acquisition. \textbf{Yunkai Zhuang:} Data curation, Investigation, Validation, Visualization. \textbf{Liye Shi:} Formal analysis, Validation, Writing -- review \& editing.

\section*{Declaration of competing interest}
The authors declare that they have no known competing financial interests or personal relationships that could have appeared to influence the work reported in this paper.

\section*{Funding}
This research was supported by Zhejiang Provincial Natural Science Foundation of China under Grant No. LQN26F020016.

\section*{Code and data availability}
Core retrieval and learning code, evaluation utilities, and summary data are available at \url{https://github.com/sungkites/LMMR}. The original benchmarks are available from their respective providers \cite{ref48,ref49,ref50}.

\section*{Declaration of generative AI and AI-assisted technologies in the writing process}
During the preparation of this work the authors used \mbox{ChatGPT} in order to improve readability and language. After using this tool, the authors reviewed and edited the content as needed and take full responsibility for the content of the publication.

\bibliographystyle{lmmr-num}
\bibliography{references}

@misc{ref1,
  author = {Park, Joon Sung and O'Brien, Joseph C. and Cai, Carrie J. and Morris, Meredith Ringel and Liang, Percy and Bernstein, Michael S.},
  title = {{Generative Agents: Interactive Simulacra of Human Behavior}},
  year = {2023},
  doi = {10.48550/arXiv.2304.03442},
  eprint = {2304.03442},
  archivePrefix = {arXiv},
  url = {https://arxiv.org/abs/2304.03442},
}

@misc{ref2,
  author = {Yao, Shunyu and Zhao, Jeffrey and Yu, Dian and Du, Nan and Shafran, Izhak and Narasimhan, Karthik and Cao, Yuan},
  title = {{ReAct: Synergizing Reasoning and Acting in Language Models}},
  year = {2022},
  doi = {10.48550/arXiv.2210.03629},
  eprint = {2210.03629},
  archivePrefix = {arXiv},
  url = {https://arxiv.org/abs/2210.03629},
}

@misc{ref3,
  author = {Shinn, Noah and Cassano, Federico and Berman, Edward and Gopinath, Ashwin and Narasimhan, Karthik and Yao, Shunyu},
  title = {{Reflexion: Language Agents with Verbal Reinforcement Learning}},
  year = {2023},
  doi = {10.48550/arXiv.2303.11366},
  eprint = {2303.11366},
  archivePrefix = {arXiv},
  url = {https://arxiv.org/abs/2303.11366},
}

@misc{ref4,
  author = {Zhong, Wanjun and Guo, Lianghong and Gao, Qiqi and Ye, He and Wang, Yanlin},
  title = {{MemoryBank: Enhancing Large Language Models with Long-Term Memory}},
  year = {2023},
  doi = {10.48550/arXiv.2305.10250},
  eprint = {2305.10250},
  archivePrefix = {arXiv},
  url = {https://arxiv.org/abs/2305.10250},
}

@misc{ref5,
  author = {Packer, Charles and Wooders, Sarah and Lin, Kevin and Fang, Vivian and Patil, Shishir G. and Stoica, Ion and Gonzalez, Joseph E.},
  title = {{MemGPT: Towards LLMs as Operating Systems}},
  year = {2023},
  doi = {10.48550/arXiv.2310.08560},
  eprint = {2310.08560},
  archivePrefix = {arXiv},
  url = {https://arxiv.org/abs/2310.08560},
}

@misc{ref6,
  author = {Xu, Wujiang and Liang, Zujie and Mei, Kai and Gao, Hang and Tan, Juntao and Zhang, Yongfeng},
  title = {{A-MEM: Agentic Memory for LLM Agents}},
  year = {2025},
  doi = {10.48550/arXiv.2502.12110},
  eprint = {2502.12110},
  archivePrefix = {arXiv},
  url = {https://arxiv.org/abs/2502.12110},
}

@misc{ref7,
  author = {Chhikara, Prateek and Khant, Dev and Aryan, Saket and Singh, Taranjeet and Yadav, Deshraj},
  title = {{Mem0: Building Production-Ready AI Agents with Scalable Long-Term Memory}},
  year = {2025},
  doi = {10.48550/arXiv.2504.19413},
  eprint = {2504.19413},
  archivePrefix = {arXiv},
  url = {https://arxiv.org/abs/2504.19413},
}

@misc{ref8,
  author = {Liu, Nelson F. and Lin, Kevin and Hewitt, John and Paranjape, Ashwin and Bevilacqua, Michele and Petroni, Fabio and Liang, Percy},
  title = {{Lost in the Middle: How Language Models Use Long Contexts}},
  year = {2023},
  doi = {10.48550/arXiv.2307.03172},
  eprint = {2307.03172},
  archivePrefix = {arXiv},
  url = {https://arxiv.org/abs/2307.03172},
}

@misc{ref9,
  author = {Bai, Yushi and Lv, Xin and Zhang, Jiajie and Lyu, Hongchang and Tang, Jiankai and Huang, Zhidian and Du, Zhengxiao and Liu, Xiao and Zeng, Aohan and Hou, Lei and Dong, Yuxiao and Tang, Jie and Li, Juanzi},
  title = {{LongBench: A Bilingual, Multitask Benchmark for Long Context Understanding}},
  year = {2023},
  doi = {10.48550/arXiv.2308.14508},
  eprint = {2308.14508},
  archivePrefix = {arXiv},
  url = {https://arxiv.org/abs/2308.14508},
}

@misc{ref11,
  author = {Hsieh, Cheng-Ping and Sun, Simeng and Kriman, Samuel and Acharya, Shantanu and Rekesh, Dima and Jia, Fei and Zhang, Yang and Ginsburg, Boris},
  title = {{RULER: What's the Real Context Size of Your Long-Context Language Models?}},
  year = {2024},
  doi = {10.48550/arXiv.2404.06654},
  eprint = {2404.06654},
  archivePrefix = {arXiv},
  url = {https://arxiv.org/abs/2404.06654},
}

@inproceedings{ref12,
  author = {Wenhan Xiong and Xiang Lorraine Li and Srinivasan Iyer and Jingfei Du and Patrick Lewis and William Wang and Yashar Mehdad and Wen-tau Yih and Sebastian Riedel and Douwe Kiela and Barlas O{\u{g}}uz},
  title = {{Answering Complex Open-Domain Questions with Multi-Hop Dense Retrieval}},
  year = {2021},
  url = {https://openreview.net/forum?id=EMHoBG0avc1},
  booktitle = {International Conference on Learning Representations},
}

@inproceedings{ref13,
  author = {Trivedi, Harsh  and
      Balasubramanian, Niranjan  and
      Khot, Tushar  and
      Sabharwal, Ashish},
  title = {{Interleaving Retrieval with Chain-of-Thought Reasoning for Knowledge-Intensive Multi-Step Questions}},
  year = {2023},
  url = {https://aclanthology.org/2023.acl-long.557/},
  booktitle = {Proceedings of the 61st Annual Meeting of the {A}ssociation for {C}omputational {L}inguistics (Volume 1: Long Papers)},
  pages = {10014--10037},
  doi = {10.18653/v1/2023.acl-long.557},
}

@misc{ref14,
  author = {Liu, Jiaqi and Su, Yaofeng and Xia, Peng and Han, Siwei and Zheng, Zeyu and Xie, Cihang and Ding, Mingyu and Yao, Huaxiu},
  title = {{SimpleMem: Efficient Lifelong Memory for LLM Agents}},
  year = {2026},
  doi = {10.48550/arXiv.2601.02553},
  eprint = {2601.02553},
  archivePrefix = {arXiv},
  url = {https://arxiv.org/abs/2601.02553},
}

@misc{ref15,
  author = {Fang, Jizhan and Deng, Xinle and Xu, Haoming and Jiang, Ziyan and Tang, Yuqi and Xu, Ziwen and Deng, Shumin and Yao, Yunzhi and Wang, Mengru and Qiao, Shuofei and Chen, Huajun and Zhang, Ningyu},
  title = {{LightMem: Lightweight and Efficient Memory-Augmented Generation}},
  year = {2025},
  doi = {10.48550/arXiv.2510.18866},
  eprint = {2510.18866},
  archivePrefix = {arXiv},
  url = {https://arxiv.org/abs/2510.18866},
}

@misc{ref16,
  author = {Zhang, Jiaquan and Zhang, Chaoning and Chen, Shuxu and Huang, Zhenzhen and Zheng, Pengcheng and Wang, Zhicheng and Guo, Ping and Mo, Fan and Bae, Sung-Ho and Zou, Jie and Wei, Jiwei and Yang, Yang},
  title = {{Lightweight LLM Agent Memory with Small Language Models}},
  year = {2026},
  doi = {10.48550/arXiv.2604.07798},
  eprint = {2604.07798},
  archivePrefix = {arXiv},
  url = {https://arxiv.org/abs/2604.07798},
}

@misc{ref18,
  author = {Yan, Sikuan and Yang, Xiufeng and Huang, Zuchao and Nie, Ercong and Ding, Zifeng and Li, Zonggen and Ma, Xiaowen and Bi, Jinhe and Kersting, Kristian and Pan, Jeff Z. and Schütze, Hinrich and Tresp, Volker and Ma, Yunpu},
  title = {{Memory-R1: Enhancing Large Language Model Agents to Manage and Utilize Memories via Reinforcement Learning}},
  year = {2025},
  doi = {10.48550/arXiv.2508.19828},
  eprint = {2508.19828},
  archivePrefix = {arXiv},
  url = {https://arxiv.org/abs/2508.19828},
}

@misc{ref19,
  author = {Wang, Yu and Takanobu, Ryuichi and Liang, Zhiqi and Mao, Yuzhen and Hu, Yuanzhe and McAuley, Julian and Wu, Xiaojian},
  title = {{Mem-$\alpha$: Learning Memory Construction via Reinforcement Learning}},
  year = {2025},
  doi = {10.48550/arXiv.2509.25911},
  eprint = {2509.25911},
  archivePrefix = {arXiv},
  url = {https://arxiv.org/abs/2509.25911},
}

@misc{ref20,
  author = {Ma, Weitao and Feng, Xiaocheng and Huang, Lei and Feng, Xiachong and Ma, Zhanyu and Xu, Jun and Gao, Jiuchong and Hao, Jinghua and He, Renqing and Qin, Bing},
  title = {{Fine-Mem: Fine-Grained Feedback Alignment for Long-Horizon Memory Management}},
  year = {2026},
  doi = {10.48550/arXiv.2601.08435},
  eprint = {2601.08435},
  archivePrefix = {arXiv},
  url = {https://arxiv.org/abs/2601.08435},
}

@misc{ref23,
  author = {Yu, Hongli and Chen, Tinghong and Feng, Jiangtao and Chen, Jiangjie and Dai, Weinan and Yu, Qiying and Zhang, Ya-Qin and Ma, Wei-Ying and Liu, Jingjing and Wang, Mingxuan and Zhou, Hao},
  title = {{MemAgent: Reshaping Long-Context LLM with Multi-Conv RL-based Memory Agent}},
  year = {2025},
  doi = {10.48550/arXiv.2507.02259},
  eprint = {2507.02259},
  archivePrefix = {arXiv},
  url = {https://arxiv.org/abs/2507.02259},
}

@misc{ref24,
  author = {Xu, Derong and Wen, Yi and Jia, Pengyue and Zhang, Yingyi and zhang, wenlin and Wang, Yichao and Guo, Huifeng and Tang, Ruiming and Zhao, Xiangyu and Chen, Enhong and Xu, Tong},
  title = {{From Single to Multi-Granularity: Toward Long-Term Memory Association and Selection of Conversational Agents}},
  year = {2025},
  eprint = {2505.19549},
  archivePrefix = {arXiv},
  url = {https://arxiv.org/abs/2505.19549},
}

@inproceedings{ref25,
  author = {Qing, Peijun  and
      Diao, Xingjian  and
      Ma, Chiyu  and
      Hassanpour, Saeed  and
      Vosoughi, Soroush},
  title = {{Tailoring Memory Granularity for Multi-Hop Reasoning over Long Contexts}},
  booktitle = {Findings of the {A}ssociation for {C}omputational {L}inguistics: {EACL} 2026},
  year = {2026},
  pages = {3648--3666},
  doi = {10.18653/v1/2026.findings-eacl.189},
  url = {https://aclanthology.org/2026.findings-eacl.189/},
}

@inproceedings{ref26,
  author = {Zeng, Qinyang  and
      Dong, Jinhao  and
      LU, Wei  and
      Du, Xiaoyong},
  title = {{{G}ra{P}-Mem: Granularity-Aware Planning for Adaptive Memory Access in Long-Horizon {LLM} Agents}},
  booktitle = {Findings of the {A}ssociation for {C}omputational {L}inguistics: {EMNLP} 2026},
  year = {2026},
  pages = {51016--51032},
  doi = {10.18653/2026.findings-emnlp.2375},
  url = {https://aclanthology.org/2026.findings-emnlp.2375/},
}

@misc{ref28,
  author = {Feng, Qi and Fan, Jicong},
  title = {{Recall What Is Missing: Composing Evidence from Long-Term Agent Memory}},
  year = {2026},
  doi = {10.48550/arXiv.2610.15642},
  eprint = {2610.15642},
  archivePrefix = {arXiv},
  url = {https://arxiv.org/abs/2610.15642},
}

@misc{ref29,
  author = {Li, Shixuan and Yang, Wei and Zhang, Peiyu and Cheng, Anzhe and Ping, Heng and Zhang, Xiaole and Thomason, Jesse and Ma, Xuezhe and Zhao, Yue and Bogdan, Paul},
  title = {{From Relevant Memories to Sufficient Evidence: Rethinking Memory Utilization in LLM Agents}},
  year = {2026},
  doi = {10.48550/arXiv.2610.22668},
  eprint = {2610.22668},
  archivePrefix = {arXiv},
  url = {https://arxiv.org/abs/2610.22668},
}

@misc{ref30,
  author = {Wei, Jiaqi and Huang, Wenxuan and Xu, Chengdong and Zhang, Xiang and Ouyang, Wanli},
  title = {{Beyond Retrieval Failure: Understanding Why LLMs Miss Evidence in Multi-Hop RAG}},
  year = {2026},
  doi = {10.48550/arXiv.2610.21184},
  eprint = {2610.21184},
  archivePrefix = {arXiv},
  url = {https://arxiv.org/abs/2610.21184},
}

@misc{ref31,
  author = {Li, Feng and Tan, Kailun and Zhou, Zhibo and Wang, Youxiang and Liu, Lejun and Li, Huaigu and Du, Fei and An, Huilong},
  title = {{KITE: Retrieval as a Decision Procedure for Long-Term Conversational Memory}},
  year = {2026},
  eprint = {2610.41521},
  archivePrefix = {arXiv},
  url = {https://arxiv.org/abs/2610.41521},
}

@misc{ref32,
  author = {Lewis, Patrick and Perez, Ethan and Piktus, Aleksandra and Petroni, Fabio and Karpukhin, Vladimir and Goyal, Naman and Küttler, Heinrich and Lewis, Mike and Yih, Wen-tau and Rocktäschel, Tim and Riedel, Sebastian and Kiela, Douwe},
  title = {{Retrieval-Augmented Generation for Knowledge-Intensive NLP Tasks}},
  year = {2020},
  doi = {10.48550/arXiv.2005.11401},
  eprint = {2005.11401},
  archivePrefix = {arXiv},
  url = {https://arxiv.org/abs/2005.11401},
}

@misc{ref33,
  author = {Karpukhin, Vladimir and O\u{g}uz, Barlas and Min, Sewon and Lewis, Patrick and Wu, Ledell and Edunov, Sergey and Chen, Danqi and Yih, Wen-tau},
  title = {{Dense Passage Retrieval for Open-Domain Question Answering}},
  year = {2020},
  doi = {10.48550/arXiv.2004.04906},
  eprint = {2004.04906},
  archivePrefix = {arXiv},
  url = {https://arxiv.org/abs/2004.04906},
}

@misc{ref34,
  author = {Khattab, Omar and Zaharia, Matei},
  title = {{ColBERT: Efficient and Effective Passage Search via Contextualized Late Interaction over BERT}},
  year = {2020},
  doi = {10.48550/arXiv.2004.12832},
  eprint = {2004.12832},
  archivePrefix = {arXiv},
  url = {https://arxiv.org/abs/2004.12832},
}

@article{ref36,
  author = {Robertson, Stephen and Zaragoza, Hugo},
  title = {{The Probabilistic Relevance Framework: BM25 and Beyond}},
  year = {2009},
  doi = {10.1561/1500000019},
  journal = {Foundations and Trends in Information Retrieval},
  volume = {3},
  number = {4},
  pages = {333--389},
  url = {https://doi.org/10.1561/1500000019},
}

@misc{ref38,
  author = {Wei, Jason and Wang, Xuezhi and Schuurmans, Dale and Bosma, Maarten and Ichter, Brian and Xia, Fei and Chi, Ed and Le, Quoc and Zhou, Denny},
  title = {{Chain-of-Thought Prompting Elicits Reasoning in Large Language Models}},
  year = {2022},
  doi = {10.48550/arXiv.2201.11903},
  eprint = {2201.11903},
  archivePrefix = {arXiv},
  url = {https://arxiv.org/abs/2201.11903},
}

@misc{ref39,
  author = {Zhou, Denny and Schärli, Nathanael and Hou, Le and Wei, Jason and Scales, Nathan and Wang, Xuezhi and Schuurmans, Dale and Cui, Claire and Bousquet, Olivier and Le, Quoc and Chi, Ed},
  title = {{Least-to-Most Prompting Enables Complex Reasoning in Large Language Models}},
  year = {2022},
  doi = {10.48550/arXiv.2205.10625},
  eprint = {2205.10625},
  archivePrefix = {arXiv},
  url = {https://arxiv.org/abs/2205.10625},
}

@inproceedings{ref40,
  author = {Carbonell, Jaime and Goldstein, Jade},
  title = {{The Use of MMR, Diversity-Based Reranking for Reordering Documents and Producing Summaries}},
  year = {1998},
  doi = {10.1145/290941.291025},
  booktitle = {Proceedings of SIGIR},
  pages = {335--336},
  url = {https://doi.org/10.1145/290941.291025},
}

@misc{ref41,
  author = {Yang, An and Li, Anfeng and Yang, Baosong and Zhang, Beichen and Hui, Binyuan and Zheng, Bo and Yu, Bowen and Gao, Chang and Huang, Chengen and Lv, Chenxu and Zheng, Chujie and Liu, Dayiheng and Zhou, Fan and Huang, Fei and Hu, Feng and Ge, Hao and Wei, Haoran and Lin, Huan and Tang, Jialong and Yang, Jian and Tu, Jianhong and Zhang, Jianwei and Yang, Jianxin and Yang, Jiaxi and Zhou, Jing and Zhou, Jingren and Lin, Junyang and Dang, Kai and Bao, Keqin and Yang, Kexin and Yu, Le and Deng, Lianghao and Li, Mei and Xue, Mingfeng and Li, Mingze and Zhang, Pei and Wang, Peng and Zhu, Qin and Men, Rui and Gao, Ruize and Liu, Shixuan and Luo, Shuang and Li, Tianhao and Tang, Tianyi and Yin, Wenbiao and Ren, Xingzhang and Wang, Xinyu and Zhang, Xinyu and Ren, Xuancheng and Fan, Yang and Su, Yang and Zhang, Yichang and Zhang, Yinger and Wan, Yu and Liu, Yuqiong and Wang, Zekun and Cui, Zeyu and Zhang, Zhenru and Zhou, Zhipeng and Qiu, Zihan},
  title = {{Qwen3 Technical Report}},
  year = {2025},
  doi = {10.48550/arXiv.2505.09388},
  eprint = {2505.09388},
  archivePrefix = {arXiv},
  url = {https://arxiv.org/abs/2505.09388},
}

@inproceedings{ref42,
  author = {Lin, Ji and Tang, Jiaming and Tang, Haotian and Yang, Shang and Chen, Wei-Ming and Wang, Wei-Chen and Xiao, Guangxuan and Dang, Xingyu and Gan, Chuang and Han, Song},
  booktitle = {Proceedings of Machine Learning and Systems},
  editor = {P. Gibbons and G. Pekhimenko and C. De Sa},
  pages = {87--100},
  title = {{AWQ: Activation-aware Weight Quantization for On-Device LLM Compression and Acceleration}},
  url = {https://proceedings.mlsys.org/paper_files/paper/2024/file/42a452cbafa9dd64e9ba4aa95cc1ef21-Paper-Conference.pdf},
  volume = {6},
  year = {2024},
}

@misc{ref45,
  author = {Maharana, Adyasha and Lee, Dong-Ho and Tulyakov, Sergey and Bansal, Mohit and Barbieri, Francesco and Fang, Yuwei},
  title = {{Evaluating Very Long-Term Conversational Memory of LLM Agents}},
  year = {2024},
  doi = {10.48550/arXiv.2402.17753},
  eprint = {2402.17753},
  archivePrefix = {arXiv},
  url = {https://arxiv.org/abs/2402.17753},
}

@misc{ref46,
  author = {Wu, Di and Wang, Hongwei and Yu, Wenhao and Zhang, Yuwei and Chang, Kai-Wei and Yu, Dong},
  title = {{LongMemEval: Benchmarking Chat Assistants on Long-Term Interactive Memory}},
  year = {2024},
  doi = {10.48550/arXiv.2410.10813},
  eprint = {2410.10813},
  archivePrefix = {arXiv},
  url = {https://arxiv.org/abs/2410.10813},
}

@misc{ref47,
  author = {Hu, Yuanzhe and Wang, Yu and McAuley, Julian},
  title = {{Evaluating Memory in LLM Agents via Incremental Multi-Turn Interactions}},
  year = {2025},
  doi = {10.48550/arXiv.2507.05257},
  eprint = {2507.05257},
  archivePrefix = {arXiv},
  url = {https://arxiv.org/abs/2507.05257},
}

@inproceedings{ref48,
  author = {Yang, Zhilin  and
      Qi, Peng  and
      Zhang, Saizheng  and
      Bengio, Yoshua  and
      Cohen, William  and
      Salakhutdinov, Ruslan  and
      Manning, Christopher D.},
  title = {{{H}otpot{QA}: A Dataset for Diverse, Explainable Multi-hop Question Answering}},
  year = {2018},
  url = {https://aclanthology.org/D18-1259/},
  booktitle = {Proceedings of the 2018 Conference on Empirical Methods in Natural Language Processing},
  pages = {2369--2380},
  doi = {10.18653/v1/D18-1259},
}

@inproceedings{ref49,
  author = {Ho, Xanh  and
      Duong Nguyen, Anh-Khoa  and
      Sugawara, Saku  and
      Aizawa, Akiko},
  title = {{Constructing A Multi-hop {QA} Dataset for Comprehensive Evaluation of Reasoning Steps}},
  year = {2020},
  doi = {10.18653/v1/2020.coling-main.580},
  booktitle = {Proceedings of the 28th International Conference on Computational Linguistics},
  pages = {6609--6625},
  url = {https://aclanthology.org/2020.coling-main.580/},
}

@article{ref50,
  author = {Trivedi, Harsh and Balasubramanian, Niranjan and Khot, Tushar and Sabharwal, Ashish},
  title = {{MuSiQue: Multihop Questions via Single-hop Question Composition}},
  year = {2022},
  doi = {10.1162/tacl_a_00475},
  journal = {Transactions of the Association for Computational Linguistics},
  volume = {10},
  pages = {539--554},
  url = {https://doi.org/10.1162/tacl_a_00475},
}

@misc{ref51,
  author = {Zhang, Yanzhao and Li, Mingxin and Long, Dingkun and Zhang, Xin and Lin, Huan and Yang, Baosong and Xie, Pengjun and Yang, An and Liu, Dayiheng and Lin, Junyang and Huang, Fei and Zhou, Jingren},
  title = {{Qwen3 Embedding: Advancing Text Embedding and Reranking Through Foundation Models}},
  year = {2025},
  doi = {10.48550/arXiv.2506.05176},
  eprint = {2506.05176},
  archivePrefix = {arXiv},
  url = {https://arxiv.org/abs/2506.05176},
}

@misc{ref52,
  author = {Hu, Edward J. and Shen, Yelong and Wallis, Phillip and Allen-Zhu, Zeyuan and Li, Yuanzhi and Wang, Shean and Wang, Lu and Chen, Weizhu},
  title = {{LoRA: Low-Rank Adaptation of Large Language Models}},
  year = {2021},
  url = {https://arxiv.org/abs/2106.09685},
  eprint = {2106.09685},
  archivePrefix = {arXiv},
}

@article{ref53,
  author = {Efron, B.},
  title = {{Bootstrap Methods: Another Look at the Jackknife}},
  year = {1979},
  doi = {10.1214/aos/1176344552},
  journal = {The Annals of Statistics},
  volume = {7},
  number = {1},
  pages = {1--26},
  url = {https://doi.org/10.1214/aos/1176344552},
}

@inproceedings{ref54,
  author = {Song, Huatong  and
      Jiang, Jinhao  and
      Tian, Wenqing  and
      Chen, Zhipeng  and
      Wu, Yuhuan  and
      Zhao, Jiahao  and
      Min, Yingqian  and
      Zhao, Wayne Xin  and
      Fang, Lei  and
      Wen, Ji-Rong},
  title = {{Smart-Searcher: Incentivizing the Dynamic Knowledge Acquisition of {LLM}s via Reinforcement Learning}},
  year = {2025},
  doi = {10.18653/v1/2025.findings-emnlp.731},
  booktitle = {Findings of the {A}ssociation for {C}omputational {L}inguistics: {EMNLP} 2025},
  pages = {13572--13586},
  url = {https://aclanthology.org/2025.findings-emnlp.731/},
}

@inproceedings{ref55,
  author = {Liu, Qichuan  and
      Zhang, Chentao  and
      Zheng, Chenfeng  and
      Hu, Guosheng  and
      Li, Xiaodong  and
      Zhang, Zhihong},
  title = {{Beyond the Answer: Advancing Multi-Hop {QA} with Fine-Grained Graph Reasoning and Evaluation}},
  year = {2025},
  doi = {10.18653/v1/2025.acl-long.1142},
  booktitle = {Proceedings of the 63rd Annual Meeting of the {A}ssociation for {C}omputational {L}inguistics (Volume 1: Long Papers)},
  pages = {23433--23456},
  url = {https://aclanthology.org/2025.acl-long.1142/},
}

@article{ref56,
  author = {Liang, Xuechen and Tao, Meiling and Xia, Yinghui and Wang, Jianhui and Li, Kun and Wang, Yijin and He, Yangfan and Yang, Jingsong and Shi, Tianyu and Wang, Yuantao and Zhang, Miao and Wang, Xueqian},
  title = {{SAGE: Self-evolving Agents with Reflective and Memory-augmented Abilities}},
  journal = {Neurocomputing},
  volume = {647},
  pages = {130470},
  year = {2025},
  doi = {10.1016/j.neucom.2025.130470},
  url = {https://doi.org/10.1016/j.neucom.2025.130470},
}

@article{ref57,
  author = {Guo, Tiezheng and Wang, Chen and Liu, Yanyi and Tang, Jiawei and Li, Pan and Xu, Sai and Yang, Qingwen and Gao, Xianlin and Li, Zhi and Wen, Yingyou},
  title = {{Leveraging inter-chunk interactions for enhanced retrieval in large language model-based question answering}},
  journal = {Neurocomputing},
  volume = {650},
  pages = {130931},
  year = {2025},
  doi = {10.1016/j.neucom.2025.130931},
  url = {https://doi.org/10.1016/j.neucom.2025.130931},
}

@inproceedings{ref58,
  title = {{{LLML}ingua-2: Data Distillation for Efficient and Faithful Task-Agnostic Prompt Compression}},
  author = {Pan, Zhuoshi  and
      Wu, Qianhui  and
      Jiang, Huiqiang  and
      Xia, Menglin  and
      Luo, Xufang  and
      Zhang, Jue  and
      Lin, Qingwei  and
      R{\"u}hle, Victor  and
      Yang, Yuqing  and
      Lin, Chin-Yew  and
      Zhao, H. Vicky  and
      Qiu, Lili  and
      Zhang, Dongmei},
  editor = {Ku, Lun-Wei  and
      Martins, Andre  and
      Srikumar, Vivek},
  booktitle = {Findings of the {A}ssociation for {C}omputational {L}inguistics: {ACL} 2024},
  month = {aug},
  year = {2024},
  address = {Bangkok, Thailand},
  publisher = {Association for Computational Linguistics},
  url = {https://aclanthology.org/2024.findings-acl.57/},
  doi = {10.18653/v1/2024.findings-acl.57},
  pages = {963--981},
}
\end{multicols}
\appendix
\renewcommand{\theHtable}{\thesection.\arabic{table}}
\renewcommand{\theHequation}{\thesection.\arabic{equation}}
\Needspace{8\baselineskip}
\section{Method and evaluation details}\label{app:settings}
\setcounter{table}{0}
\renewcommand{\thetable}{A\arabic{table}}
\setcounter{equation}{0}
\renewcommand{\theequation}{A\arabic{equation}}
\subsection{Model configuration and input templates}
Tables~\ref{tab:settings} and \ref{tab:modelsettings} give the retrieval parameters and answer-model configurations. Table~\ref{tab:inputtemplates} gives the input templates; Table~\ref{tab:format} compares answer-format instructions.
\begin{table}[H]
\centering
\caption{Method and optimization parameters.}\label{tab:settings}
\small
\setlength{\tabcolsep}{3.5pt}
\renewcommand{\arraystretch}{1.06}
\begin{tabularx}{\linewidth}{@{}L{.16\linewidth}L{.30\linewidth}Y@{}}
\toprule
Category & Parameter & Value \\
\midrule
Memory selection & $B$; $K_e$; $K_a$; $K_s$ & 400 tokens; 2; 2; 3 \\
Scoring & $\tau$; $\alpha$; $\beta$; $\eta$ & 2; 1.2; 1; 0.7 \\
Node lengths & Coarse / fine & 192 / 96 reranker tokens \\
Training examples & Training / development; coarse--fine pairs & 1,600 / 64; 1,206 paired-positive questions \\
Controls and negatives & Coarse-only / matched coarse-only; hard negatives & 381 / 1,600 questions; top 2 non-supporting negatives per granularity \\
Loss weights & $m$; $\lambda_r$; $\lambda_B$; $\lambda_c$ & 0.5; 0.5; 0.25; 0.1 \\
LoRA & Base model; rank / scaling & Qwen3-Reranker 0.6B; 8 / 16 \\
Adaptation & Projections; trainable parameters & Attention $q$ and $v$ in 28 layers; 1,146,880 \\
Optimization & Optimizer; learning rate; weight decay & AdamW; 5e-5; 0.01 \\
Updates & Questions per step; steps; gradient-norm limit & 4; 600; 1 \\
Precision & Initialization; training; evaluation & 8-bit GGUF dequantization; BF16 base + FP32 adapters; merged FP16 \\
Candidate pools & HotpotQA; 2Wiki; MuSiQue & 10 / 20 / 30 / 50; 10; 20 paragraphs \\
Pool expansion & Additional HotpotQA sources & BM25 (question + initial anchor), title deduplication \\
Budget analysis & Procedure & 192 questions; selected evidence clipped at source boundaries \\
Answer decoding & Quantization; temperature; maximum output & Q4\_K\_M; 0; 128 tokens \\
Prompt and execution & Full reference-prompt limit; devices & 768 tokens; A100 retrieval and Orin answering \\
Intervals and tests & Paired resampling; primary-test correction & Support-source sets; 4,000 draws; Holm \\
Compression & Model; retained fraction & LLMLingua-2, XLM-R Large; 0.75 \\
\bottomrule
\end{tabularx}
\end{table}

\begin{table}[H]
\centering
\caption{Configurations of the eight answer models.}\label{tab:modelsettings}
\small
\setlength{\tabcolsep}{3.5pt}
\renewcommand{\arraystretch}{1.06}
\begin{tabularx}{\linewidth}{@{}L{.16\linewidth}YL{.14\linewidth}@{}}
\toprule
Provider & Model identifier & Quantization \\
\midrule
Alibaba & Qwen/Qwen3.5-0.8B & Q4\_K\_M \\
Google & google/gemma-3-1b-it & Q4\_K\_M \\
Tencent & tencent/Hunyuan-1.8B-Instruct & Q4\_K\_M \\
Alibaba & Qwen/Qwen3.5-2B & Q4\_K\_M \\
Hugging Face & HuggingFaceTB/SmolLM3-3B & Q4\_K\_M \\
IBM & ibm-granite/granite-4.0-micro & Q4\_K\_M \\
Microsoft & microsoft/Phi-4-mini-instruct & Q4\_K\_M \\
Alibaba & Qwen/Qwen3.5-4B & Q4\_K\_M \\
\bottomrule
\end{tabularx}
\end{table}

\begin{table}[H]
\centering
\caption{Model instructions and input fields used in retrieval and answering.}\label{tab:inputtemplates}
\small
\setlength{\tabcolsep}{3.5pt}
\renewcommand{\arraystretch}{1.06}
\begin{tabularx}{\linewidth}{@{}L{.22\linewidth}Y@{}}
\toprule
Component & Instruction or input \\
\midrule
Relevance instruction & Given a question, retrieve relevant passages that answer the question. \\
Conditional instruction & Given the question and the known evidence, retrieve a different passage that supplies the missing information needed to answer the question. Prefer complementary evidence over repetition of the known evidence. \\
Conditional fields & \texttt{<Query>: \{question\}}\newline
\texttt{Known evidence: \{anchor\}}\newline
\texttt{<Document>: \{candidate\}}\newline
Decision tokens: \texttt{yes} / \texttt{no}. \\
Answer instruction & Answer using only the supplied evidence. Return a concise direct answer without explanation. If the evidence is insufficient, say "not enough information". \\
Answer fields & \texttt{Evidence: \{selected evidence\}}\newline
\texttt{Question: \{question\}}\newline
\texttt{Answer:} \\
\bottomrule
\end{tabularx}
\end{table}

The matched coarse-only model uses the same 1,600 question groups, order, backbone, LoRA configuration, optimizer, and 600 updates as LMMR. It uses coarse positives and the corresponding hard and redundant negatives. A supported coarse node from the same source is used when a training positive crosses a coarse-node boundary.

LLMLingua-2 uses the official XLM-R Large checkpoint with a retained fraction of 0.75, targeting roughly a quarter less input. This produces comparable evidence lengths: 186.42 tokens for the compressed matched retriever and 185.32 for LMMR on HotpotQA. Context filtering is disabled, token selection follows the official defaults, and the same answer-model tokenizer measures all input lengths.

\subsection{Evidence metrics}
Supporting-source recall is the proportion of annotated supporting sources represented in the selected evidence, averaged over questions. Support density is the mean per-question fraction of evidence tokens belonging to annotated support text; source labels and titles remain in the denominator. For HotpotQA and 2WikiMultihopQA, strict support completeness requires every annotated support sentence to be retained in full. For MuSiQue, it records whether all annotated supporting paragraphs are represented in the selected evidence.

Answer-in-context rate (AIC) checks whether the normalized canonical answer string occurs in the evidence, excluding \texttt{yes}, \texttt{no}, and \texttt{noanswer} items. Answer F1 and EM use normalized short-answer scoring. Table~\ref{tab:aliases} additionally reports MuSiQue scores maximized over the official answer and its aliases.

\subsection{Normalized MMR and statistical comparisons}
Normalized MMR \cite{ref40} combines the QE score with directional redundancy:
\begin{equation}
\widetilde{s}_{\mathrm{QE}}(d)=
\frac{s_{\mathrm{QE}}(d)-\min s_{\mathrm{QE}}}
{\max s_{\mathrm{QE}}-\min s_{\mathrm{QE}}+\epsilon},
\label{eq:mmrnorm}
\end{equation}
\begin{equation}
s_{\mathrm{MMR}}(d)=\eta\widetilde{s}_{\mathrm{QE}}(d)-(1-\eta)R(a,d).
\label{eq:mmr}
\end{equation}
The minimum and maximum are computed over the current anchor's eligible coarse candidates.

Intervals use 4,000 paired bootstrap draws. Questions with identical complete sets of annotated support sources are resampled together. The 21 primary F1 comparisons across the three datasets use Holm correction, including the matched retriever and the two learned-compression controls. Question-type intervals are pointwise; the score-weight and fixed-evidence budget sweeps are descriptive.

\Needspace{8\baselineskip}
\section{Supplementary results}\label{app:results}
\setcounter{table}{0}
\renewcommand{\thetable}{B\arabic{table}}
\setcounter{figure}{0}
\renewcommand{\thefigure}{B\arabic{figure}}
\renewcommand{\theHfigure}{B.\arabic{figure}}
\subsection{Interference and memory composition}
\begin{figure}[H]
\centering\includegraphics[width=0.85939\linewidth]{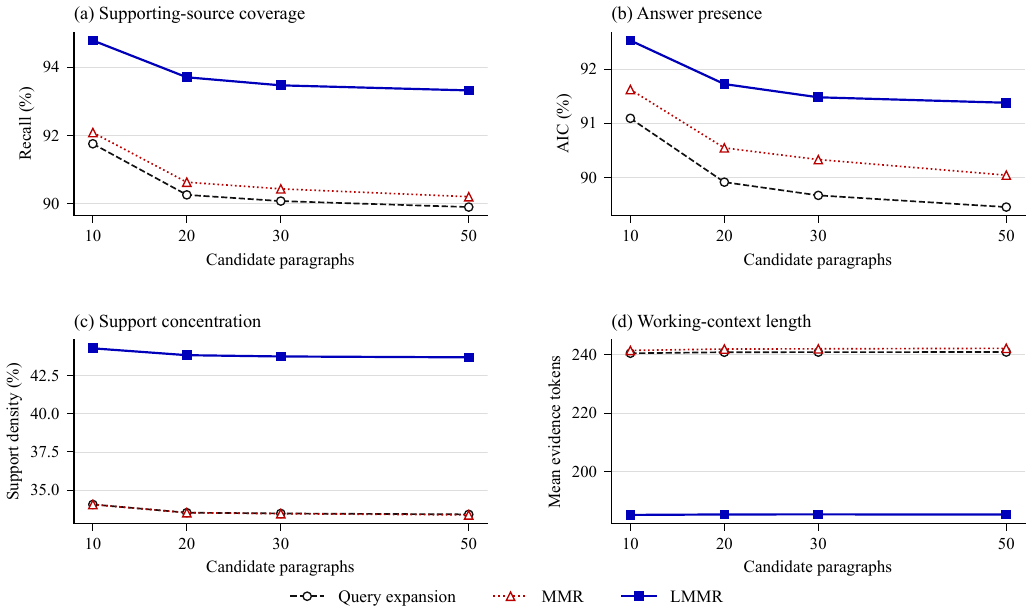}
\caption{Memory behavior as the HotpotQA pool grows from 10 to 50 paragraphs. Each condition uses 7,405 validation questions and the same 400-token budget. Panels report supporting-source recall, non-boolean answer-in-context rate (AIC), mean per-question support density, and mean evidence length. Markers show the evaluated settings.}\label{fig:app-pool}
\end{figure}

At 50 candidate paragraphs, LMMR retains 93.32\% supporting-source recall and 43.69\% support density, compared with 89.89\% and 33.41\% for QE, while using 185.24 rather than 240.95 tokens (Fig.~\ref{fig:app-pool}). Table~\ref{tab:pool} gives the corresponding F1 and EM scores.
\begin{table}[H]
\centering
\caption{Candidate-pool analysis on 7,405 HotpotQA validation questions.}\label{tab:pool}
\small
\setlength{\tabcolsep}{3.5pt}
\renewcommand{\arraystretch}{1.13}
\begin{tabularx}{\linewidth}{@{}Yrrrrrrrr@{}}
\toprule
\multirow{2}{*}{Method} & \multicolumn{2}{c}{10 paragraphs} & \multicolumn{2}{c}{20 paragraphs} & \multicolumn{2}{c}{30 paragraphs} & \multicolumn{2}{c}{50 paragraphs} \\
\cmidrule(lr){2-3}\cmidrule(lr){4-5}\cmidrule(lr){6-7}\cmidrule(lr){8-9}
& F1 $\uparrow$ & EM $\uparrow$ & F1 $\uparrow$ & EM $\uparrow$ & F1 $\uparrow$ & EM $\uparrow$ & F1 $\uparrow$ & EM $\uparrow$ \\
\midrule
Query expansion & 59.24 & 44.04 & 58.56 & 43.55 & 58.42 & 43.42 & 58.27 & 43.27 \\
Normalized MMR & 59.55 & 44.27 & 58.97 & 43.86 & 58.79 & 43.69 & 58.61 & 43.50 \\
Coarse-only adaptation & 60.90 & 45.36 & 60.14 & 44.79 & 60.04 & 44.71 & 59.91 & 44.63 \\
FH & 57.20 & 42.39 & 56.72 & 42.12 & 56.65 & 42.08 & 56.55 & 41.99 \\
\textbf{LMMR} & \textbf{62.05} & \textbf{46.40} & \textbf{61.61} & \textbf{46.10} & \textbf{61.47} & \textbf{45.96} & \textbf{61.42} & \textbf{45.87} \\
\bottomrule
\end{tabularx}
\end{table}

\subsection{Score weight and evidence budget}
\begin{table}[H]
\centering
\caption{Conditional-score weight sensitivity on 7,405 HotpotQA validation questions.}\label{tab:alpha}
\small
\setlength{\tabcolsep}{3.5pt}
\renewcommand{\arraystretch}{1.13}
\begin{tabularx}{.52\linewidth}{@{}Yrrr@{}}
\toprule
$\alpha$ & F1 $\uparrow$ & EM $\uparrow$ & Tokens $\downarrow$ \\
\midrule
0.8 & \textbf{61.55} & \textbf{45.96} & 185.95 \\
1.0 & 61.47 & 45.94 & 185.60 \\
1.2 & 61.47 & \textbf{45.96} & 185.32 \\
1.5 & 61.46 & 45.94 & 185.02 \\
1.8 & 61.44 & 45.91 & \textbf{185.00} \\
\bottomrule
\end{tabularx}
\end{table}

\begin{table}[H]
\centering
\caption{Fixed-evidence budget analysis on 192 questions.}\label{tab:budget}
\small
\setlength{\tabcolsep}{3.5pt}
\renewcommand{\arraystretch}{1.13}
\begin{tabularx}{\linewidth}{@{}Yrrrrrrrrr@{}}
\toprule
\multirow{2}{*}{Budget} & \multicolumn{3}{c}{Query expansion} & \multicolumn{3}{c}{MMR} & \multicolumn{3}{c}{LMMR} \\
\cmidrule(lr){2-4}\cmidrule(lr){5-7}\cmidrule(lr){8-10}
& F1 $\uparrow$ & EM $\uparrow$ & Tokens $\downarrow$ & F1 $\uparrow$ & EM $\uparrow$ & Tokens $\downarrow$ & F1 $\uparrow$ & EM $\uparrow$ & Tokens $\downarrow$ \\
\midrule
140 & 36.21 & 24.48 & 134.76 & 36.83 & 23.96 & 135.11 & \textbf{39.97} & \textbf{27.08} & \textbf{129.08} \\
180 & 41.81 & 27.08 & 168.58 & 42.50 & 26.56 & 169.69 & \textbf{46.65} & \textbf{30.21} & \textbf{153.44} \\
220 & 49.12 & 34.90 & 196.84 & 48.39 & 32.81 & 198.45 & \textbf{55.63} & \textbf{36.98} & \textbf{169.94} \\
280 & 53.98 & 38.54 & 221.94 & 53.27 & 36.46 & 224.48 & \textbf{58.90} & \textbf{39.06} & \textbf{179.64} \\
400 & 54.83 & \textbf{40.10} & 232.34 & 54.11 & 38.02 & 234.97 & \textbf{59.37} & 39.06 & \textbf{183.03} \\
\bottomrule
\end{tabularx}
\end{table}

\subsection{Question type and reasoning depth}
\begin{figure}[H]
\centering\includegraphics[width=0.96323\linewidth]{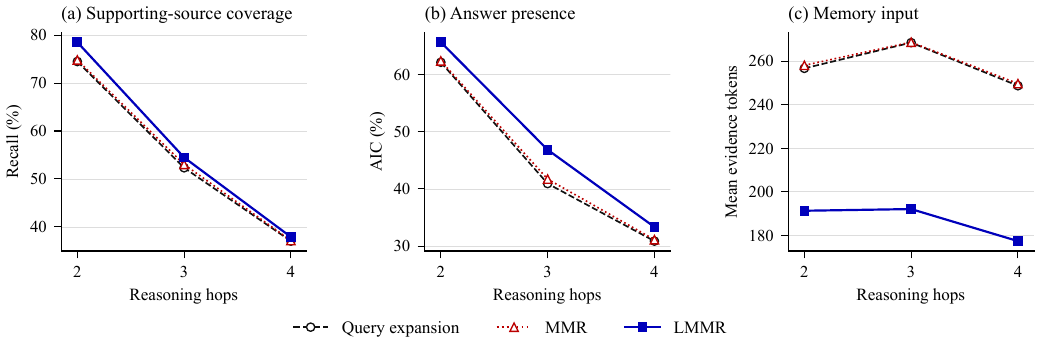}
\caption{Supporting-source recall, answer presence (AIC), and mean evidence length by MuSiQue reasoning depth. The two-, three-, and four-hop groups contain 1,252, 760, and 405 questions, respectively. All groups use the same 400-token evidence budget and frozen 2B answer model. Lines connect the observed hop groups.}\label{fig:app-hops}
\end{figure}

Across MuSiQue's two-, three-, and four-hop groups, LMMR reduces mean evidence input by 25.5\%, 28.5\%, and 28.7\% relative to QE. Supporting-source recall and answer presence remain higher in each group (Fig.~\ref{fig:app-hops}). Table~\ref{tab:types} provides the answer scores by question type.
\begin{table}[H]
\centering
\caption{Answer F1 and EM by dataset and question type.}\label{tab:types}
\small
\setlength{\tabcolsep}{3.5pt}
\renewcommand{\arraystretch}{1.13}
\begin{tabularx}{\linewidth}{@{}Yrrrrr@{}}
\toprule
\multirow{2}{*}{Dataset and type} & \multirow{2}{*}{$n$} & \multicolumn{2}{c}{F1 $\uparrow$} & \multicolumn{2}{c}{EM $\uparrow$} \\
\cmidrule(lr){3-4}\cmidrule(lr){5-6}
& & QE & LMMR & QE & LMMR \\
\midrule
2Wiki bridge comparison & 2751 & \textbf{42.44} & 41.49 & \textbf{40.31} & 38.71 \\
2Wiki comparison & 3040 & 48.43 & \textbf{49.32} & 42.34 & \textbf{42.57} \\
2Wiki compositional & 5236 & 39.32 & \textbf{43.72} & 27.12 & \textbf{29.76} \\
\addlinespace[3pt]
2Wiki inference & 1549 & 30.50 & \textbf{32.97} & 10.78 & \textbf{12.07} \\
Hotpot bridge & 5918 & 62.62 & \textbf{66.41} & 46.54 & \textbf{49.78} \\
\addlinespace[3pt]
Hotpot comparison & 1487 & 41.67 & \textbf{41.83} & \textbf{31.00} & 30.73 \\
MuSiQue 2hop & 1252 & 35.47 & \textbf{40.97} & 24.44 & \textbf{28.27} \\
MuSiQue 3hop & 760 & 17.72 & \textbf{22.63} & 9.61 & \textbf{13.55} \\
MuSiQue 4hop & 405 & 12.40 & \textbf{13.36} & \textbf{8.89} & 8.15 \\
\bottomrule
\end{tabularx}
\end{table}

\subsection{Scoring and answer-format variants}
\begin{table}[H]
\centering
\caption{Single-reference and alias-aware scoring on MuSiQue.}\label{tab:aliases}
\small
\setlength{\tabcolsep}{3.5pt}
\renewcommand{\arraystretch}{1.13}
\begin{tabularx}{\linewidth}{@{}L{.28\linewidth}*{4}{>{\centering\arraybackslash}X}@{}}
\toprule
\multirow{2}{*}{Method} & \multicolumn{2}{c}{Single reference} & \multicolumn{2}{c}{Alias-aware} \\
\cmidrule(lr){2-3}\cmidrule(lr){4-5}
& F1 $\uparrow$ & EM $\uparrow$ & F1 $\uparrow$ & EM $\uparrow$ \\
\midrule
Query expansion & 26.02 & 17.17 & 28.02 & 20.07 \\
Normalized MMR & 25.97 & 17.13 & 27.93 & 20.02 \\
Coarse-only adaptation & 26.65 & 17.50 & 28.64 & 20.31 \\
FH & 26.05 & 16.84 & 27.94 & 19.74 \\
\textbf{LMMR} & \textbf{30.58} & \textbf{20.27} & \textbf{32.57} & \textbf{23.50} \\
\bottomrule
\end{tabularx}
\end{table}

\begin{table}[H]
\centering
\caption{HotpotQA with a uniform short-answer instruction: 7,405 questions and 10 candidate paragraphs.}\label{tab:format}
\small
\setlength{\tabcolsep}{3.5pt}
\renewcommand{\arraystretch}{1.13}
\begin{tabularx}{.61\linewidth}{@{}Yrr@{}}
\toprule
Method & F1 $\uparrow$ & EM $\uparrow$ \\
\midrule
Query expansion & 65.13 & 50.83 \\
Normalized MMR & 65.47 & 51.10 \\
Coarse-only adaptation & 66.77 & 52.13 \\
FH & 63.08 & 49.32 \\
\textbf{LMMR} & \textbf{67.92} & \textbf{53.30} \\
\bottomrule
\end{tabularx}
\end{table}

\subsection{Paired statistical comparisons}
\begin{table}[H]
\centering
\caption{Primary F1 differences, 95\% paired-bootstrap intervals, and Holm-adjusted p-values across 21 comparisons. Positive differences favor LMMR.}\label{tab:statistics}
\small
\setlength{\tabcolsep}{3.5pt}
\renewcommand{\arraystretch}{1.08}
\begin{tabularx}{.97\linewidth}{@{}Yrlr@{}}
\toprule
Baseline & F1 difference & 95\% interval & Holm p \\
\midrule
\multicolumn{4}{l}{\textit{2WikiMultihopQA}} \\
Query expansion & +2.14 & [1.55, 2.72] & 0.0052 \\
Normalized MMR & +1.77 & [1.20, 2.35] & 0.0052 \\
Coarse-only adaptation & +2.28 & [1.69, 2.86] & 0.0052 \\
FH & +5.05 & [4.52, 5.57] & 0.0052 \\
Matched coarse-only & +0.55 & [0.05, 1.08] & 0.1102 \\
QE + LLMLingua-2 & +5.72 & [5.04, 6.42] & 0.0052 \\
Matched coarse + LLMLingua-2 & +3.99 & [3.34, 4.61] & 0.0052 \\
\midrule
\multicolumn{4}{l}{\textit{HotpotQA}} \\
Query expansion & +3.06 & [2.43, 3.70] & 0.0052 \\
Normalized MMR & +2.68 & [2.08, 3.34] & 0.0052 \\
Coarse-only adaptation & +1.44 & [0.84, 2.04] & 0.0052 \\
FH & +4.82 & [4.24, 5.44] & 0.0052 \\
Matched coarse-only & +0.22 & [-0.31, 0.75] & 0.4146 \\
QE + LLMLingua-2 & +5.57 & [4.77, 6.40] & 0.0052 \\
Matched coarse + LLMLingua-2 & +2.76 & [2.05, 3.45] & 0.0052 \\
\midrule
\multicolumn{4}{l}{\textit{MuSiQue}} \\
Query expansion & +4.55 & [3.28, 5.83] & 0.0052 \\
Normalized MMR & +4.61 & [3.33, 5.90] & 0.0052 \\
Coarse-only adaptation & +3.93 & [2.75, 5.18] & 0.0052 \\
FH & +4.53 & [3.40, 5.71] & 0.0052 \\
Matched coarse-only & +2.08 & [0.90, 3.22] & 0.0052 \\
QE + LLMLingua-2 & +4.19 & [2.80, 5.64] & 0.0052 \\
Matched coarse + LLMLingua-2 & +1.35 & [0.04, 2.66] & 0.1102 \\
\bottomrule
\end{tabularx}
\par\smallskip
\begin{minipage}{.97\linewidth}
\footnotesize P-values use the plus-one finite-draw correction with 4,000 bootstrap draws. Holm correction is applied jointly to the 21 F1 comparisons.
\end{minipage}
\end{table}

\subsection{Additional model and retrieval controls}\label{app:nemotron}
\begin{table}[H]
\centering
\caption{Zero-shot transfer to the full validation sets of 2WikiMultihopQA and MuSiQue. MuSiQue uses single-reference scoring.}\label{tab:external}
\small
\setlength{\tabcolsep}{3.5pt}
\renewcommand{\arraystretch}{1.13}
\begin{tabularx}{\linewidth}{@{}Yrrrrrr@{}}
\toprule
\multirow{2}{*}{Method} & \multicolumn{3}{c}{2WikiMultihopQA (12,576 questions)} & \multicolumn{3}{c}{MuSiQue (2,417 questions)} \\
\cmidrule(lr){2-4}\cmidrule(lr){5-7}
& F1 $\uparrow$ & EM $\uparrow$ & Tokens $\downarrow$ & F1 $\uparrow$ & EM $\uparrow$ & Tokens $\downarrow$ \\
\midrule
Query expansion & 41.12 & 31.67 & 242.47 & 26.02 & 17.17 & 259.01 \\
Normalized MMR & 41.49 & 31.99 & 243.72 & 25.97 & 17.13 & 259.96 \\
Coarse-only adaptation & 40.98 & 31.46 & 244.10 & 26.65 & 17.50 & 255.13 \\
FH & 38.22 & 28.97 & \textbf{173.23} & 26.05 & 16.84 & \textbf{188.03} \\
\addlinespace[3pt]
\textbf{LMMR} & \textbf{43.26} & \textbf{32.63} & 177.94 & \textbf{30.58} & \textbf{20.27} & 189.21 \\
\bottomrule
\end{tabularx}
\end{table}

\begin{table}[H]
\centering
\caption{Additional Nemotron Nano 4B results: raw-output scoring on 7,405 HotpotQA questions with Q4\_K\_M quantization.}\label{tab:nemotron}
\small
\setlength{\tabcolsep}{3.5pt}
\renewcommand{\arraystretch}{1.06}
\begin{tabularx}{.80\linewidth}{@{}Yrrrrrr@{}}
\toprule
Answer model & QE F1 & QE EM & MMR F1 & MMR EM & LMMR F1 & LMMR EM \\
\midrule
Nemotron Nano 4B & 8.23 & 0.30 & 8.28 & 0.30 & 8.48 & 0.45 \\
\bottomrule
\end{tabularx}
\end{table}

\Needspace{40\baselineskip}
\section{Published results under their original settings}\label{app:published}
\setcounter{table}{0}
\renewcommand{\thetable}{C\arabic{table}}
Table~\ref{tab:published} lists IRCoT \cite{ref13}, Smart-Searcher \cite{ref54}, and PER-QA \cite{ref55} results with their answer models, retrieval settings, and evaluation subsets. Overall results and question-type subsets appear in separate blocks. PER-QA uses GPT-4o for planning and Llama3.1-8B-Instruct for answering.
\begin{table}[H]
\centering
\caption{Published results and LMMR reference results under the indicated settings.}\label{tab:published}
\small
\setlength{\tabcolsep}{3.5pt}
\renewcommand{\arraystretch}{1.08}
\begin{tabularx}{\linewidth}{@{}L{.17\linewidth}L{.20\linewidth}Yrr@{}}
\toprule
Method & Answer model & Evaluation setting & EM & F1 \\
\midrule
\multicolumn{5}{l}{\textit{HotpotQA: overall evaluation sets}} \\
IRCoT~\cite{ref13} & Flan-T5-XL 3B & Open domain / 500 & --- & 56.70 \\
IRCoT~\cite{ref13} & Flan-T5-XXL 11B & Open domain / 500 & --- & 59.10 \\
IRCoT~\cite{ref13} & GPT-3 175B & Open domain / 500 & 49.30 & 60.70 \\
Smart-Searcher~\cite{ref54} & Qwen2.5-7B & Open domain / 500 & --- & 59.00 \\
PER-QA RAG~\cite{ref55} & 8B + GPT-4o planner & Processed set / overall & --- & 62.90 \\
\textbf{LMMR} & Qwen3.5-2B & Controlled pool / 7,405 & 45.96 & 61.47 \\
\textbf{LMMR} & Qwen3.5-4B & Controlled pool / 7,405 & 57.18 & 71.76 \\
\midrule
\multicolumn{5}{l}{\textit{HotpotQA: question-type subsets}} \\
IRCoT~\cite{ref13} & GPT-3 175B & Open domain / bridge & 45.80 & 58.50 \\
PER-QA RAG~\cite{ref55} & 8B + GPT-4o planner & Processed 2-hop bridge / 500 & --- & 47.70 \\
\textbf{LMMR} & Qwen3.5-2B & Controlled pool / 5,918 bridge & 49.78 & 66.41 \\
\midrule
\multicolumn{5}{l}{\textit{2WikiMultihopQA: overall evaluation sets}} \\
IRCoT~\cite{ref13} & Flan-T5-XL 3B & Open domain / 500 & --- & 55.10 \\
IRCoT~\cite{ref13} & Flan-T5-XXL 11B & Open domain / 500 & --- & 66.50 \\
IRCoT~\cite{ref13} & GPT-3 175B & Open domain / 500 & 57.70 & 68.00 \\
Smart-Searcher~\cite{ref54} & Qwen2.5-7B & Open domain / 500 & --- & 61.20 \\
PER-QA RAG~\cite{ref55} & 8B + GPT-4o planner & Processed set / overall & --- & 54.60 \\
\textbf{LMMR} & Qwen3.5-2B & Controlled pool / 12,576 & 32.63 & 43.26 \\
\midrule
\multicolumn{5}{l}{\textit{2WikiMultihopQA: question-type subsets}} \\
PER-QA RAG~\cite{ref55} & 8B + GPT-4o planner & Processed 2-hop bridge / 500 & --- & 42.90 \\
\textbf{LMMR} & Qwen3.5-2B & Controlled pool / 5,236 compositional & 29.76 & 43.72 \\
\midrule
\multicolumn{5}{l}{\textit{MuSiQue: overall evaluation sets}} \\
IRCoT~\cite{ref13} & Flan-T5-XL 3B & Open domain / 2--4 hops / 500 & --- & 32.30 \\
IRCoT~\cite{ref13} & Flan-T5-XXL 11B & Open domain / 2--4 hops / 500 & --- & 30.80 \\
IRCoT~\cite{ref13} & GPT-3 175B & Open domain / 2--4 hops / 500 & 26.50 & 36.50 \\
Smart-Searcher~\cite{ref54} & Qwen2.5-7B & Open domain / mixed hops / 500 & --- & 33.80 \\
PER-QA RAG~\cite{ref55} & 8B + GPT-4o planner & Processed set / 2--4 hops & --- & 24.30 \\
\textbf{LMMR} & Qwen3.5-2B & Controlled pool / 2,417 & 20.27 & 30.58 \\
\midrule
\multicolumn{5}{l}{\textit{MuSiQue: question-type subsets}} \\
IRCoT~\cite{ref13} & GPT-3 175B & Open domain / 2-hop & 34.20 & 43.80 \\
PER-QA RAG~\cite{ref55} & 8B + GPT-4o planner & Processed 2-hop bridge / 500 & --- & 30.60 \\
\textbf{LMMR} & Qwen3.5-2B & Original 20 paragraphs / 1,252 2-hop & 28.27 & 40.97 \\
\bottomrule
\end{tabularx}
\end{table}

\end{document}